\documentclass[a4paper,12pt]{article}
\usepackage[margin=2.5cm]{geometry}
\usepackage{dsfont}
\usepackage{graphicx} % Required for inserting images
\usepackage[colorlinks=true,allcolors=magenta]{hyperref}
\usepackage{amsmath}
\usepackage[nameinlink]{cleveref}
\usepackage{natbib}

\usepackage[scaled=0.92]{helvet}

\usepackage{setspace}
\title{Deep learning of geometrical cell division rules}
\author{Alexandre Durrmeyer, Jean-Christophe Palauqui$^*$, Philippe Andrey$^*$}
\date{Université Paris-Saclay, INRAE, AgroParisTech,\\ Institute Jean-Pierre Bourgin for Plant Sciences (IJPB),\\ 78000, Versailles, France\\}

\begin{document}
\maketitle
\noindent
$^*$ Corresponding authors:\newline
\texttt{jean-christophe.palauqui@inrae.fr}\newline
\texttt{philippe.andrey@inrae.fr}
\cleardoublepage
\section*{Abstract}
The positioning of new cellular walls during cell division plays a key role in shaping plant tissue organization. 
The influence of cell geometry on the positioning of division planes has been previously captured into various geometrical rules. Accordingly, linking cell shape to division orientation has relied on the comparison between observed division patterns and predictions under specific rules. The need to define \textit{a priori} the tested rules is a fundamental limitation of this hypothesis-driven approach. As an alternative, we introduce a data-based approach to investigate the relation between cell geometry and  division plane positioning, exploiting the ability of deep neural network to learn complex relationships across multidimensional spaces. Adopting an image-based cell representation, we show how division patterns can be learned and predicted from mother cell geometry using a UNet architecture modified to operate on cell masks. Using synthetic data and \textit{A. thaliana} embryo  cells, we  evaluate the model performances on a wide range of diverse cell shapes and division patterns. We find that the trained model  accounted for embryo division patterns  that were previously  irreconcilable under existing geometrical rules. 
Our work shows the potential of deep networks to understand cell division patterns and to generate new hypotheses on the control of cell division positioning.

\vspace{1cm}
\noindent
\textbf{Keywords}: division plane positioning; cell shape; modeling; UNet; \textit{Arabidopsis thaliana}.
   
\newpage

\section{Introduction}

Morphogenesis is a complex phenomenon during which organ shape emerges as a result of individual cell behaviors and interactions.
In plant development and morphogenesis, where semi-rigid cellular walls forbid cell rearrangements \citep{Cosgrove_2005}, cell division is key to tissue organization.
The positioning (location and orientation) of the new wall formed during a division determines the formation of new cell layers and reshapes the geometrical and topological properties of the surrounding tissue \citep{Sahlin_2010}. These properties in turn affect cell growth, tissue mechanics, cell-to-cell communication, and future divisions \citep{Alim_2012}. Understanding the determinants and regulations of cell wall positioning during division is thus essential for a better understanding of plant morphogenesis.

Multiple factors have been related to division plane positioning \citep{Livanos_2019}. The influence of cell geometry has been captured in so-called geometrical division rules \citep{Hofmeister_1863,Sachs_1887}. For instance, Errera's shortest path rule stipulates that equal division (i.e., producing two daughter cells of equal volumes) is oriented so as to minimize the contact area between the daughter cells \citep{Errera_1888}.
Later work by \citet{Minc_2011} used microscopic wells to constrain sea urchin zygote shapes and showed that cell geometry does influence  division plane positioning. 
Additional cues have been proposed to play a role in division orientation.
For instance, external mechanical constraints were shown to influence the division orientation of cells \citep{Lintilhac_1984}. In \textit{Arabidopsis thaliana} shoot apical meristem (SAM), it has been shown that divisions tend to align with the direction of maximal tensile stress \citep{Louveaux_2016}.
Genetic control has been revealed by studies on auxin signaling mutants with altered cell division behaviors \citep{Yoshida_2014, Livanos_2019, Vaddepalli_2021}. 
Topological properties were also linked to cell division. For instance, it was proposed that  global topological properties such as cell centrality were emerging from, and maintained by, the division regime in \textit{A. thaliana} SAM \citep{Jackson_2019}. Furthermore, \citet{Matz_2022} showed that the orientation of the division plane could be better predicted when combining topological and geometrical features  than when using geometrical features alone.

Most studies often use the shortest wall rule, or its three-dimensional extension \citep{Martinez_2018}, as a baseline to identify non-geometrical influences on cell division \citep{Yoshida_2014,Louveaux_2016,Jackson_2019}.
Accordingly, the area minimization rule is generally considered a default behavior overridden by additional mechanisms. 
However, the failure of a given geometrical rule to account for observed division patterns does not necessarily imply the absence of a relation between cell geometry and division plane positioning. 
For example, a new geometric division rule in which the division plane is the minimal surface passing through the cell center was recently evaluated \citep{Moukhtar_2019}. This rule successfully predicted the initial divisions in \textit{A. thaliana} embryo up to the 16-cell stage and most of the division patterns leading to the 32-cell stage \citep{Laruelle_2022}. Remarkably, the rule could predict unequal divisions at the 8C-16C transition \citep{Moukhtar_2019} that previous geometric modeling could not predict \citep{Yoshida_2014}. The key difference between the two studies was in modeling division planes as either perfectly planar \citep{Yoshida_2014} or  curved surfaces \citep{Moukhtar_2019}. Their diverging conclusions on the 8C-16C division illustrate a fundamental limit of the hypothesis-driven approach, in which \textit{a priori} given geometric rules are assessed.
The question then arises whether it is possible to assess the existence of geometrical principles linking division plane positioning to cell shape without \textit{a priori} designed rules.

A data-driven approach based on cell shape with no geometrical \textit{a priori} on the cell geometry/division plane relation is a promising way to address the limitations of the hypothesis-driven approach. 
Geometrical division rules can be considered as complex mappings between cell geometries and division planes.
As universal approximators of functions between high-dimensional input and output spaces \citep{Hornik_1989}, deep neural networks (DNNs) are ideal candidates to  map such  relationships \citep{Chai_2024}. 
Importantly, unlike rule-based approaches relying on human-selected geometrical features, neural networks can autonomously extract relevant features from training samples \citep{Turley_2024, Hallou_2021}. 
For instance, \citet{Zinchenko_2023} could accurately classify cells using the geometric features extracted from a neural network. 
Hence, we hypothesize an artificial neural network could potentially learn to reproduce any geometrical division rule when such a relation exists in the training data.

Despite the potential of deep learning to assess the existence of geometrical division rules, several issues must be addressed to turn our proposition into a concrete methodology. 
The first question is to define the model input and output spaces, based on an appropriate formalism to represent cells and their divisions. 
The second question is to design an appropriate architecture and learning strategy to train the network. 
The third question is to evaluate the relevance of the performed training. 
This involves assessing the accuracy of the predictions but also ensuring that a generic geometric rule has indeed been learned during the training of the neural network.

In this paper, we address these questions and demonstrate the ability of deep neural networks to link the positioning of division planes to the geometry of mother cells.
First, we introduce an image-based formalism for learning division plane positioning, casting the prediction of a division plane as a partitioning problem that can be addressed with a U-net network architecture.
Then, we modify the U-Net to better fit the specificity of this problem and validate the network predictions on controlled synthetic data to establish its potential and limitations. 
Finally, we evaluate the network on observed divisions from \textit{A. thaliana} embryo cells from the 1C to 32C stages, covering situations where previous works highlighted possibly contrasted situations on the existence of geometrical rules.

\section{Results}

\subsection{Turning division plane prediction into a CNN-tractable problem}

\begin{figure}
\centering\includegraphics[width=\textwidth]{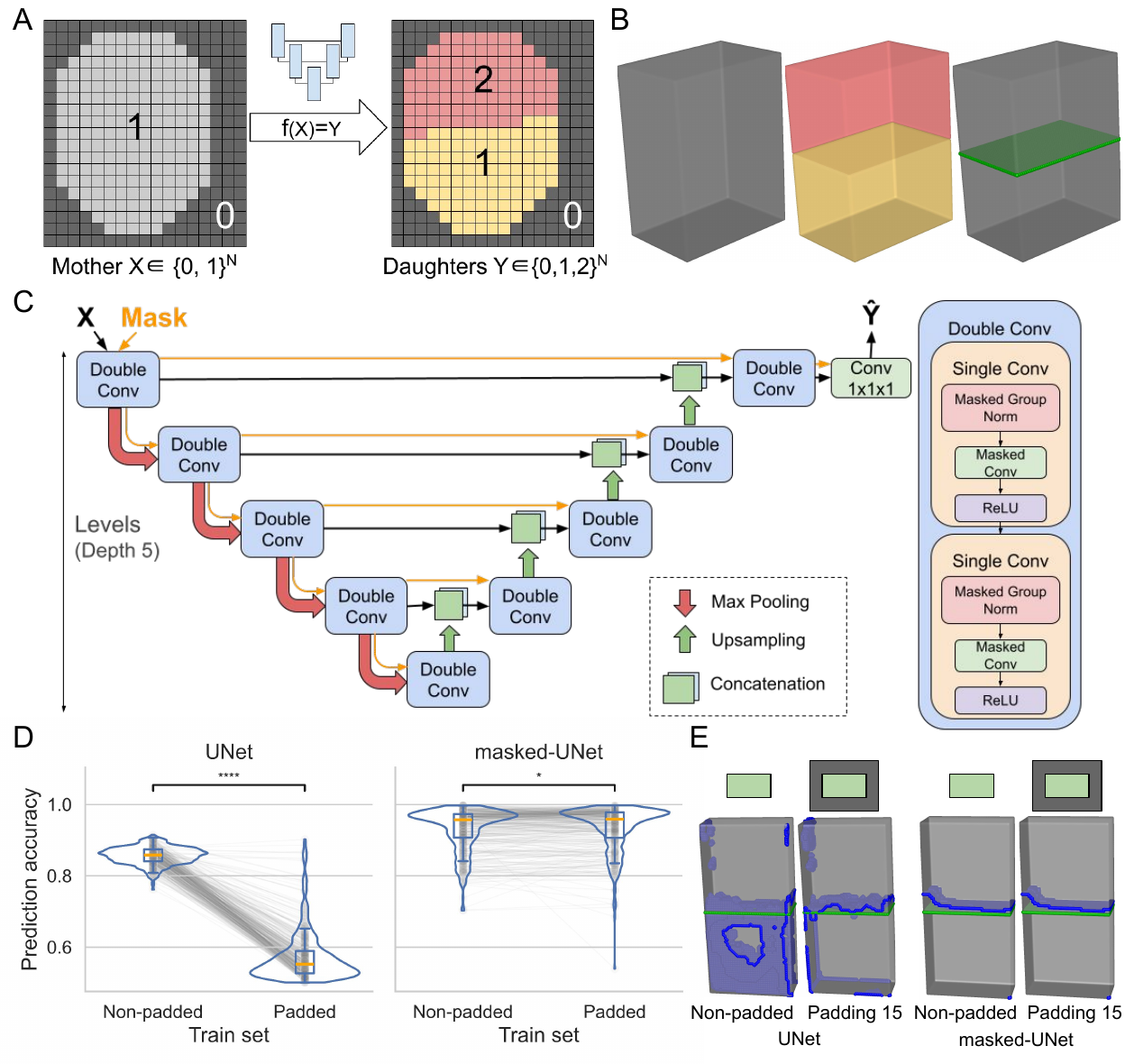}
\caption[Proposed CNN-based approach to learn 3D cell division patterns.]{Proposed CNN-based approach to learn 3D cell division patterns.
(A)~Casting the prediction of cell division positioning as a partitioning problem on a 3D lattice (here illustrated in 2D). A division rule is represented by a CNN-encoded function $f$ mapping $X$ (mother cell) to $Y$ (daughter cells). 
(B)~A 3D cuboid cell before and after division (3D mesh representation). \textit{Left}: Mother cell before division. \textit{Middle}: Daughter cells after division. \textit{Right}: Division plane between the two daughter cells (\textit{Green}). 
(C)~Architecture of the masked-UNet. A mask is propagated through the model to restrain the output of all network operations.
(D)~Effect of random padding on accuracy for UNet and masked-UNet. Wilcoxon test: $P$=3.06e-67 (\textit{Left}) and $P$=0.044 (\textit{Right}) (N=400). 
(E)~Predictions in a same shape with UNet and masked-UNet under different paddings. 
}
\label{fig:initialModifications}
\end{figure}

Segmented microscopy images are the primary data when studying cell division patterns. Given the success of convolutional neural networks (CNN) in low-level image-processing tasks, we adopted an image-based approach to represent cells and their divisions in 3D.
Rather than predicting the division plane itself, we formalized cell division as a discrete partitioning problem where the binary mask of the mother cell is partitioned into two regions corresponding to the daughter cells \citep{Moukhtar_2019}. 
Predicting a division pattern thus consists in labeling each voxel of the mother cell as either 1 or 2 (\Cref{fig:initialModifications}A).
In this formalism, the division plane is implicitly defined as the interface between the two daughter labels (\Cref{fig:initialModifications}B). There is no assumption or restriction on the division interface, which can deviate from perfectly planar geometry to account for observed curvature in division planes.
Note that our formalism assumes no size and shape change of the mother cell before and after division. This is a biologically plausible assumption for plants and other species with cellular walls, as well as in animal cells were early embryogenesis is dominated by cell cleavages in spite of possible temporary shape changes during the division itself.

We propose to use a UNet-3D neural network to learn functions (division rules) partitioning mother cells into their daughter cells. Indeed, our representation of cell division establishes a strong parallel between the prediction of a division pattern and the segmentation of an image into labeled regions.
The UNet-3D is a convolutional neural network architecture known in biomedical imaging for its efficiency in image segmentation tasks, even with small datasets \citep{Ronneberger_2015,Cicek_2016,Falk_2018}.

To evaluate the feasibility of our approach, we trained a model to reproduce synthetic division patterns. The training dataset consisted of cuboidal mother cells and their divisions in two daughter cells according to Errera's rule (symmetric division minimizing the contact area between the two daughter cells; \citealt{Errera_1888}) (\Cref{fig:initialModifications}B). 
Mother cells were generated with diverse volumes, elongations, and flatnesses (see Material and Methods, ``Synthetic cells'').
The training loss successfully decreased and converged (\Cref{fig:sup-initialTraining}A). Together with the evolution of prediction during training (\Cref{fig:sup-initialTraining}C), this highlighted the capacity of the UNet-3D to progressively learn features of Errera's rule in cuboids. 
The accuracy of the predicted partitioning (proportion of correctly labeled voxels in the predicted division compared to the target pattern) on the training shapes was generally over 95\% (\Cref{fig:sup-initialTraining}B), a high value compared with accuracy obtained on moderately altered division patterns (\Cref{fig:sup-accuracy-erosion-synthetic}). This  showed good agreement between target and predicted planes, as corroborated by visual examination (\Cref{fig:sup-initialTraining}D). These results showed the UNet-3D capacity to learn Errera's rule on cuboid shapes.

\subsection{A modified UNet architecture to learn cell division patterns}

Since we aim to predict division plane positioning in a cell-autonomous context, the predicted plane should exclusively depend on mother cell geometry. This implies that the network training and predictions should be invariant or equivariant regarding the ordering of the daughter labels, the orientation of the mother cell, and its position in the image.

As expected, the native UNet was sensitive to the consistency of daughter labels between predicted and target patterns during training (\Cref{fig:sup-label-invariance}). We enforced insensitivity to label permutation by defining a label-symmetric cross-entropy, retaining the smallest cross-entropy loss value obtained by comparing each prediction to the target with and without permutation of the daughter labels (see Material and Methods). The same permutation was used in the computation of prediction accuracy. By construction, this symmetrization was sufficient to ensure invariance to the ordering of daughter labels (\Cref{fig:sup-label-invariance}B).

As cells in the training set had their axes aligned with the image axes, the UNet was strongly sensitive to rotation applied at the inference stage (\Cref{fig:sup-rotation-invariance}).
To enforce robustness to shape orientation, we randomly rotated the images during training, a common data augmentation technique \citep{Quiroga_2019}.
As expected, this greatly reduced the prediction sensitivity to rotation  (\Cref{fig:sup-rotation-invariance}). 

To evaluate position invariance, we compared model predictions under different padding sizes around the input shapes and found a strong sensitivity to padding (\Cref{fig:initialModifications}DE, \textit{Left}). 
This is consistent with previous reports showing that CNNs can extract information from convolution artefacts at image borders \citep{Kayhan_2020}(\Cref{fig:sup-propagatedMask}A). 
We modified the original UNet architecture and introduced a ``masked-UNet'', in which the mother cell mask was propagated through the network to restrict computations to the cell interior (\Cref{fig:initialModifications}C). The mask was applied at each level of the masked-UNet and in each channel, ensuring that activations outside the mother cell remained null and did not affect activations inside the cell (\Cref{fig:sup-propagatedMask}B; see Material and Methods). Comparing the masked-UNet predictions against the original model confirmed that the sensitivity to padding was almost completely abolished (\Cref{fig:initialModifications}DE, \textit{Right}). 
Altogether, the implemented modifications allowed the masked-UNet to exhibit translation, rotation, and label ordering invariance or equivariance. 

\subsection{The masked-UNet can learn different division rules}

To further characterize the generic capacities of the masked-UNet, we trained it on different shapes and division rules.
Four experimental conditions were considered, by combining two shape categories (cuboids and ellipsoids) with two division rules: Errera's rule (symmetric division following the shortest path) and ``Anti-Hertwig's rule'' (symmetric division orthogonal to the smallest axis), a custom rule orthogonal to  Hertwig's rule  \citep{Hertwig_1884} (\Cref{fig:syntheticTraining}A). A distinct model was trained for each of these conditions.

The losses for both training and validation sets decreased and converged towards similar plateaus, confirming the masked-UNet learning ability with similar dynamics for diverse shapes and division rules (\Cref{fig:syntheticTraining}B).
The accuracy distributions over the training sets confirmed that the network could accurately fit the training data (\Cref{fig:syntheticTraining}C). The accuracy distributions on the corresponding test sets and 3D visualizations of predictions confirmed that the networks could generalize equally well the learned rules onto previously unseen shapes (\Cref{fig:syntheticTraining}CD). 
These results suggest the masked-UNet can capture arbitrary division principles in arbitrary cell geometries.

\begin{figure}
\includegraphics[width=\linewidth]{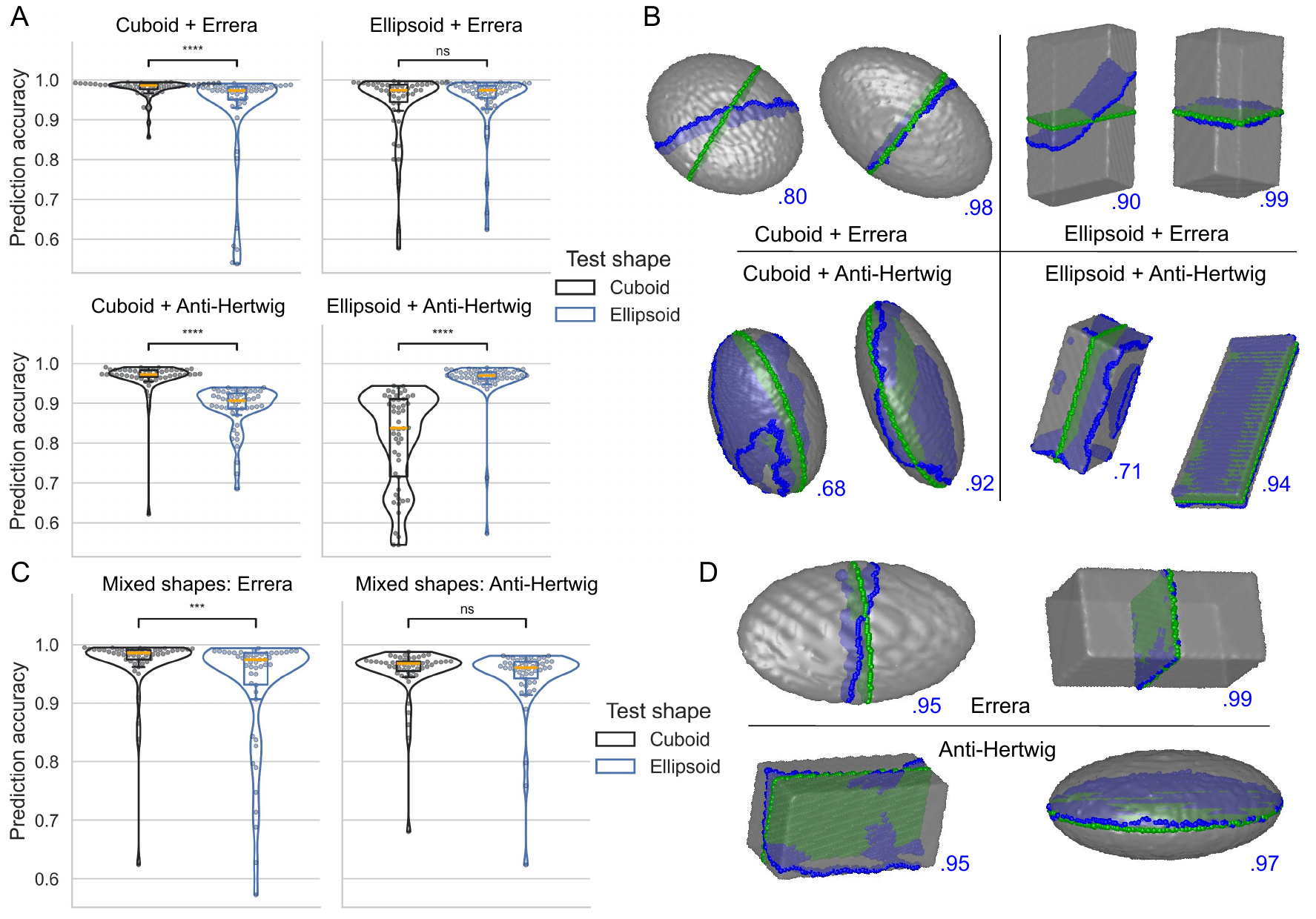}
\caption{Learning and generalization between different shape categories.
(A)~Accuracy distribution of trained models on both shape categories. The model training conditions (shape and division rule) are indicated above each graph. Unpaired Wilcoxon test (N=50), $P$=5.59e-05 (Cuboid+Errera), 0.58 (Ellipsoid+Errera), 3.57e-15 (Cuboid+Anti-Hertwig), 1.17e-15 (Ellipsoid+Anti-Hertwig). 
(B)~Predictions in shapes from the alternative  category. The values give prediction accuracy.
(C)~Accuracy distribution for models trained on mixtures of shapes. $P$=0.003 (Errera) and 0.12 (Anti-Hertwig) (N=50).
(D)~Predictions of models trained on mixtures of shapes.
\textit{Green}: Target division plane.
\textit{Blue}: Predicted division plane. 
}
\label{fig:mixedSynthetic}
\end{figure}

A key question is whether the four trained networks learned a generic geometrical relation between cell shape and plane positioning or a relation specific to each shape category (cuboids or ellipsoids). We thus examined how the networks generalized between shape categories.
For Errera's division rule, the model trained on ellipsoids generalized with high accuracy on cuboids, and conversely (\Cref{fig:mixedSynthetic}AB). On the contrary, the generalization of the Anti-Hertwig rule to the alternative shape resulted on average in a decreased accuracy (\Cref{fig:mixedSynthetic}A). The decrease was more pronounced for the model trained on ellipsoids than for the model trained on cuboids, suggesting that the predictions with the two models relied on different features with asymmetric cross-generalization potential. Visualizing predictions in cuboids revealed that the model trained on ellipsoids predicted division planes starting from two cuboid corners, passing through the cuboid center, and ending through two opposite corners (\Cref{fig:mixedSynthetic}B,  \textit{Bottom right}). This suggests the model had learned to join the two points of maximum curvature while passing through the geometrical center. 
Overall, these cross-shape generalization experiments are consistent with the hypothesis that different shape-plane relations have been captured by models trained on different shapes.

We then wondered if a shared geometrical principle could be learned by a single model. For each of the two division rules, we separately trained and tested a network over a mixture of cuboids and ellipsoids. In both cases, the accuracy of the mixed model was similar or only slightly lower compared to shape-specific models (\Cref{fig:mixedSynthetic}CD). Importantly, it was systematically more accurate than the specialized models tested on the alternative shape (compare with \Cref{fig:mixedSynthetic}A). 
This experiment confirmed that a single model could capture a common geometrical rule over a set of heterogeneous shapes. 

\subsection{The masked-UNet is limited to well-defined divisions}\label{sec:ambiguous}

\begin{figure}
\includegraphics[width=\linewidth]{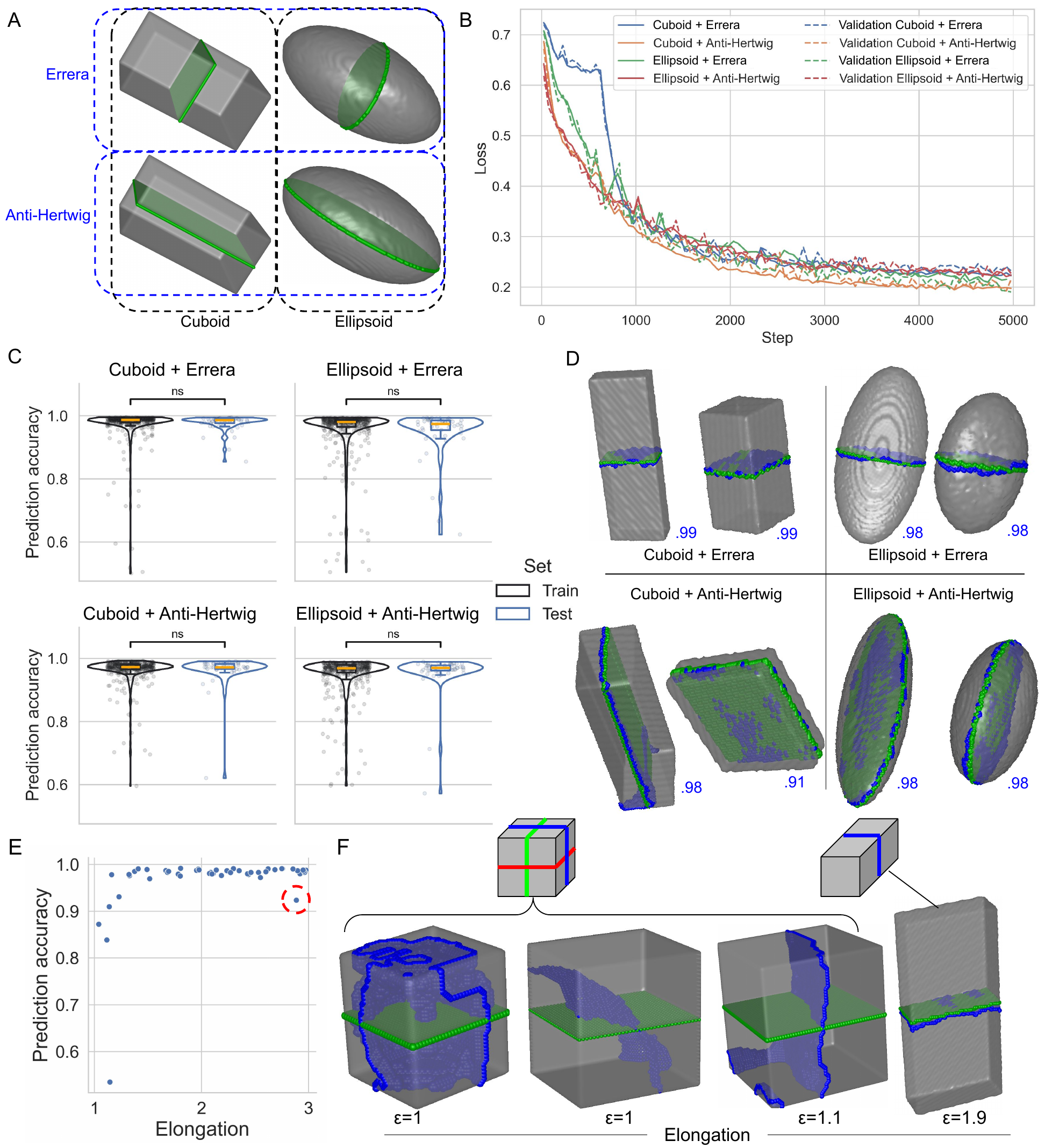}
\caption[Learning division rules on synthetic shapes.]{Learning division rules on synthetic shapes.
(A)~Examples from the four synthetic datasets. 
(B)~Training and validation losses during training on each dataset.
(C)~Accuracy distributions on training (\textit{Black}; N=400) and test (\textit{Blue}; N=50) sets. Unpaired Wilcoxon test, $P$=0.34 (Cuboid+Errera), 0.17 (Ellipsoid+Errera), 0.64 (Cuboid+Anti-Hertwig), and 0.52 (Ellipsoid+Anti-Hertwig).
(D)~Representative test set predictions. \textit{Blue}: predicted division plane; \textit{Green}: target division plane.  The values are the prediction accuracy.
(E)~Cuboid+Errera model: accuracy distribution in the test set as a function of cuboid elongation (N=50). An outlier with high elongation but low accuracy is circled in red.
(F)~Cuboid+Errera model: predictions in cuboids with different elongations. $\epsilon$: shape elongation. The first three shapes are ill-defined cases (low-elongation cuboids) with several divisions satisfying equally well, or almost equally well, Errera's rule. The fourth shape is unambiguous (elongated cuboid),  with a unique solution satisfying Errera's rule.
}
\label{fig:syntheticTraining}
\end{figure}

Most previous experiments pointed to a few inaccurate predictions (see, e.g., distributions tails in \Cref{fig:syntheticTraining}C). 
Analyzing the predictions of the model trained on cuboids divided according to Errera's rule revealed a marked drop in accuracy for cuboids with low elongation (\Cref{fig:syntheticTraining}E). These cases correspond to ambiguous situations, in which division planes with different orientations can satisfy equally well the geometrical rule. Division of such``ambiguous'' mother cells is thus ill-defined in the sense that it is not unique. For example, cuboids with low elongation display minimal surfaces orthogonal to both the first and second principal axes. Though some predictions in such shapes displayed accurate or plausible alternative divisions, others exhibited a mixture between alternative planes (\Cref{fig:syntheticTraining}F). This suggests the model was subject to a numerical indetermination in which extracted features related to concurring division planes were competing with one another.

\Cref{fig:syntheticTraining}E exhibited an outlier with pronounced elongation but lower accuracy than similarly elongated shapes (circled in red in \Cref{fig:syntheticTraining}E). This altered prediction was due to a specific orientation, as revealed by systematically rotating this shape: the division was accurately predicted at most rotation angles   but there was a marked drop in accuracy at one particular orientation (\Cref{fig:sup-rotation-ambiguity}). This can be interpreted by the deterministic and arbitrary left/right assignment of daughter labels. Since a cuboid shape is invariant by rotation of 180$^o$ around any of its axes, there is necessarily an angle where a label switch occurs.

Overall, these observations point to a limitation of the model: due to its deterministic inferences and to the training protocol penalizing any valid alternative division, the model cannot select in a stochastic manner among a set of strictly equivalent, alternative partitions and generally makes incorrect prediction in case of concurring solutions. 

\subsection{The masked-UNet can learn Errera's division rule on real cells}

In contrast to synthetic data, biological data are composed of complex, variable shapes and are limited in the number of available samples. 
We thus evaluated if the masked-UNet could learn a synthetic division rule in real cell shapes.

\begin{figure}
\includegraphics[width=\textwidth]{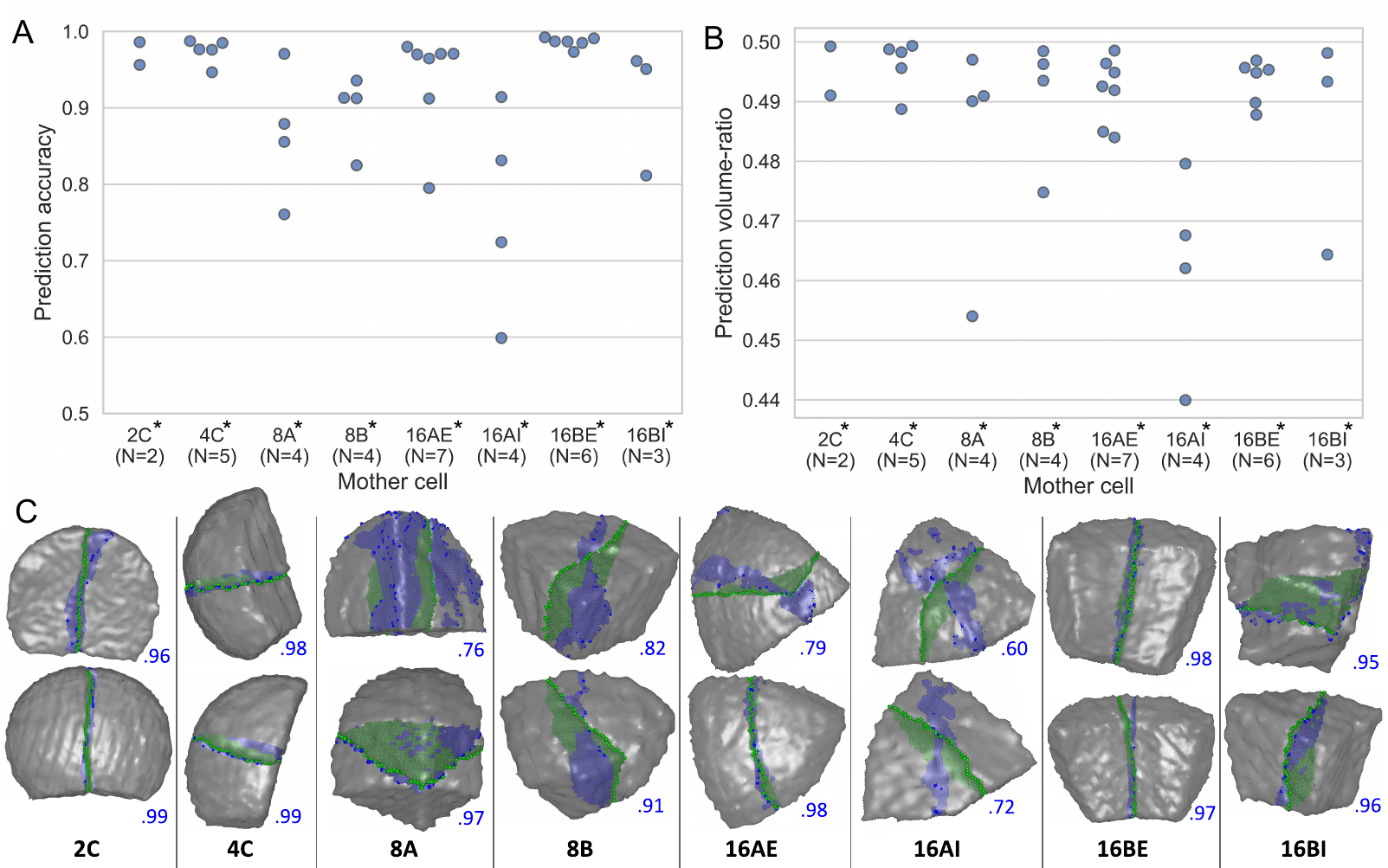}
\caption{Learning Errera's rule in real cell shapes. Ground-truth patterns were generated by simulating numerically the rule in \textit{A. thaliana} embryo cells from stages 2C to 16C and from all embryo domains (A = apical; B = basal; E = external; I = internal). 
(A)~Prediction accuracy on the test set. 
(B)~Prediction volume-ratio.
(C)~Predictions at the different generations and in the different embryo domains (two example cells are shown in each case). The number associated with each pattern gives the prediction accuracy.
\textit{Green:} simulated Errera's division plane (target); \textit{Blue:} predicted plane.}
\label{fig:ErreraEmbryo}
\end{figure}

We trained a model on \textit{A. thaliana} early embryo cells that we had artificially divided in two daughters of equal volume according to Errera's rule (see Material and Methods). 
The cells were taken at various embryo stages (2C to 16C) and domains (apical/basal, external/internal).
The trained model could faithfully reproduce Errera's rule in most cells (\Cref{fig:ErreraEmbryo}AC). 
The predicted volume-ratios ranged close to 0.5, indicating that the model had correctly captured the symmetry of the division rule (\Cref{fig:ErreraEmbryo}B). Hence, the large prediction errors observed in some cases (\Cref{fig:ErreraEmbryo}A) could be attributed more to an error in the predicted position of the division plane rather than to the relative volumes of the daughter cells. 
Predictions in 8A and 16AI domains were generally incorrect, as in few cells from 8B, 16AE, and 16BI domains. Remarkably, Errera's rule displayed variable division patterns across cells in these domains (\Cref{fig:ErreraEmbryo}C, \textit{Green} planes). 
This suggests that cell shapes with incorrect predictions could represent ill-defined cases for Errera's rule, with several, nearly equivalent division patterns. 

Overall, these results show that the model can learn and predict a geometrical division rule from variable real cell shapes, especially in well-defined situations.

\subsection{The masked-UNet can learn real division patterns}\label{sec:individual-training}

In \textit{A. thaliana} embryo, the first four generations of cell divisions (of stages 1C to 8C) are stereotyped. At the same time, they show a large variety of cell shapes and division patterns between generations and embryonic domains (basal/apical, external/internal), with different plane orientations (anticlinal or periclinal) and division volume-ratios (equal, slightly unequal, and strongly unequal)  \citep{Yoshida_2014,Moukhtar_2019,Laruelle_2022} (\Cref{fig:sup-embryo-development} and \Cref{fig:sup-generation-properties}B). It was shown previously that these stages could be recapitulated by a common ``centroid rule'' (plane of minimum area passing by the cell center) \citep{Moukhtar_2019}, which encompasses Errera's rule in the symmetric case (divisions at 1C and 2C). As a first assessment of the capacity of masked-UNet to learn real division patterns, we  examined whether it could learn and predict separately each of these stereotyped divisions. 

\begin{figure}
\includegraphics[width=\textwidth]{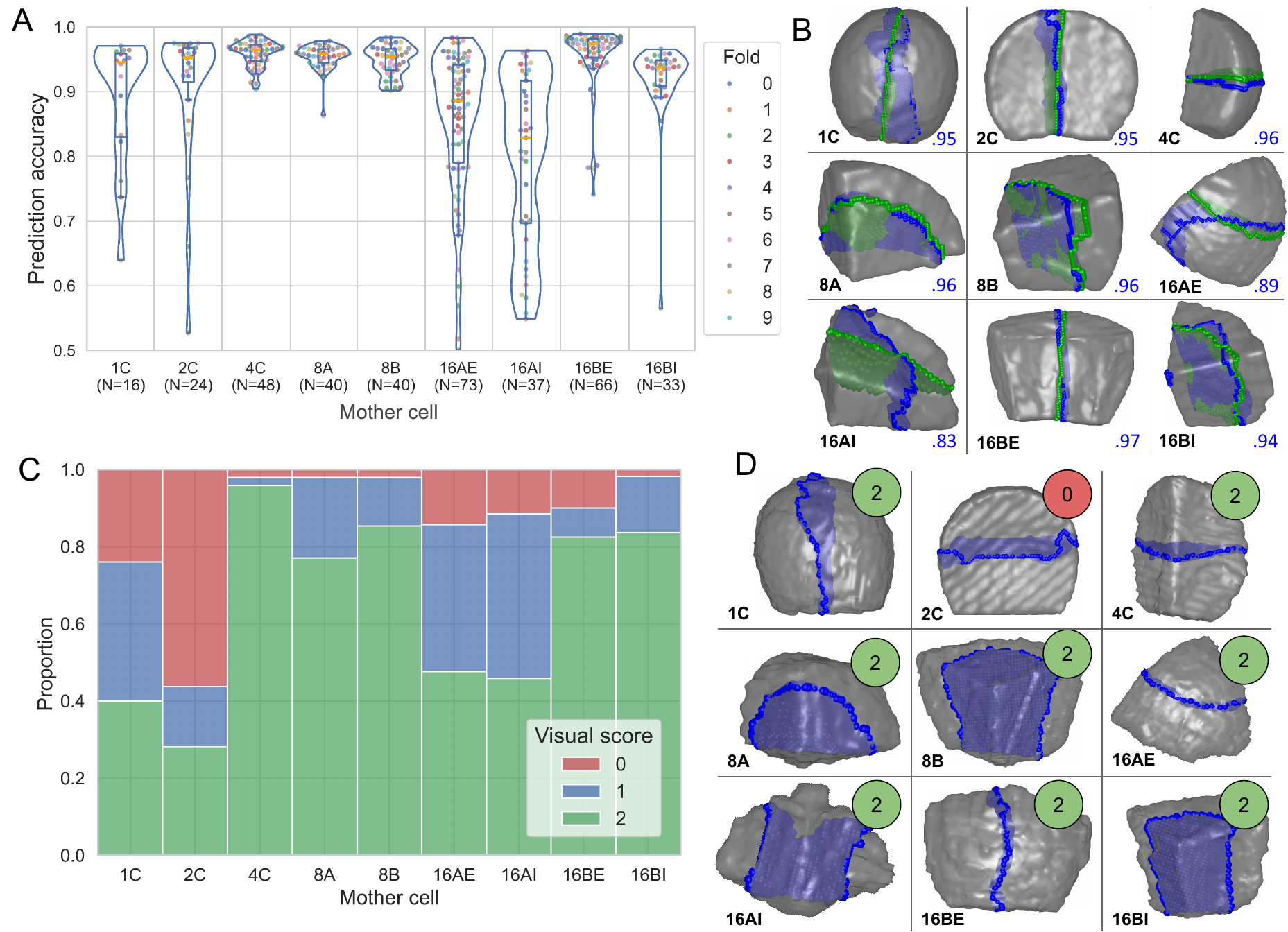}
\caption{Per-generation and per-domain learning of observed division patterns in \textit{Arabidopsis thaliana} embryo. A 10-fold approach was used to train and test distinct models for each generation and embryo domain between 1C and 16C stages.
(A)~Prediction accuracy pooled over the 10 folds. 
(B)~Model predictions in reconstructed mother cells (merged daughters). \textit{Green}: target (observed) division plane. \textit{Blue}: predicted division plane. The value associated with each pattern is the prediction accuracy.
(C)~Visual assessment of predictions in native (non-divided) cells. \textit{0}: incorrect prediction; \textit{1}: noisy or partially correct prediction; \textit{2}:  correct prediction. 
(D)~Predictions in native cells.  Corresponding numbers are the visually attributed scores. For each generation and domain, the displayed cell illustrates the most common  score.
}
\label{fig:individualCells}
\end{figure}

Since embryos were fixed for imaging, the training sets were composed of pairs of sister cells (target patterns) and the corresponding mother cells (input shapes) reconstructed by merging these sister cells (see Material and Methods). 
Given the limited number of available cells (total of 373, see \Cref{tab:embryo-cells}), we used a $k$-fold approach \citep{Zhang_2023a} to evaluate the model on a sufficient number of test cells without reducing drastically the training set size. We trained $k=10$ models, each time leaving out a different 10\% of the dataset for testing. The same data augmentation was applied as for synthetic data (random rotation only).
Since the per-fold average accuracy was highly reproducible (\Cref{fig:sup-kfoldMeans}), 
we pooled test set predictions over all folds. 
The prediction accuracy was high (\Cref{fig:individualCells}A), well above the accuracy of slightly altered target patterns (\Cref{fig:sup-accuracy-erosion-embryo}), indicating that the models had successfully captured the geometrical relationships between cell geometries and division planes for each division between 1C and 8C stages (\Cref{fig:individualCells}B). 

The high accuracy at the 1C stage was unexpected given the infinite number of equivalent planes that can divide the axially symmetric cell shapes of this stage. 
We hypothesized that reconstructing mother cells by merging their daughters could leave a merging scar at the cell surface  (\Cref{fig:sup-mergingScar}) or could encompass a geometric deformation of the mother cell shape due to the division. We tested whether the network took advantage of these cues. For each stage and each domain, the best-performing model over the 10 folds was evaluated on native (not yet divided)  cells of the corresponding stage and domain. Since they were not reconstructed by merging daughters, these native cells did not contain division cues.
Because no groundtruth was available, we relied on the stereotypical nature of the division patterns to visually rate predictions as 0 (incorrect), 1 (plausible orientation or moderate departure from expected pattern), or 2 (correct) (\Cref{fig:sup-noMerging}A).
Many predictions in native 1C and 2C cells were not correct (\Cref{fig:individualCells}C), contrasting with predictions in the corresponding reconstructed mother cells (\Cref{fig:individualCells}A). This confirmed that merging cues were probably exploited for predictions in these cells.
Since reconstructed 1C cells showed no global shape alteration compared with native cells (\Cref{fig:sup-noMerging}BC), the merging scar was likely the primary cue informing the predictions.
Reconstructed 2C cell shapes exhibited higher elongation and lower flatness compared with native ones (\Cref{fig:sup-noMerging}BC), suggesting the merging cue could be a deformation induced by the division wall.
All other stages displayed consistent predictions between native and merged cells (\Cref{fig:individualCells}C), showing that, for most division patterns, the network was not relying on reconstruction cues.  

We next extended our analysis to the fifth generation (divisions of 16C cells), which corresponds to a peak in mother cell shape variability over early embryo stages \citep{Laruelle_2022}. Observed divisions at this stage are contrasted in their orientations and volume-ratios \citep{Yoshida_2014,Laruelle_2022} (\Cref{fig:sup-embryo-development} and \Cref{fig:sup-generation-properties}B). 
Divisions in the basal domain are stereotyped, with anticlinal, equal divisions in the external domain (16BE), and periclinal, unequal divisions in the internal domain (16BI). The apical domain exhibits variable division patterns. In the external domain (16AE), divisions are anticlinal with three different orientations and are slightly unequal. In the internal domain (16AI), three main orientations and diverse volume-ratios are observed. 
It was shown previously that the stereotyped 16BE and variable 16AE divisions could be explained deterministically based on the same ``centroid'' rule as for stages 1C to 8C (see above), which was also the case for some orientations in the 16AI case; on the contrary, the 16BI division was not reproduced by this rule \citep{Laruelle_2022}.

The 16BE divisions were the most accurately predicted (\Cref{fig:individualCells}A), in accordance with their invariant orientation and volume-ratio, and with their agreement with the centroid rule \citep{Laruelle_2022}.
Most 16AE divisions were also correctly predicted. However, several cells exhibited predicted planes that mixed different orientations observed experimentally (\Cref{fig:individualCells}B). These patterns, reminiscent of predictions in low-elongation cuboids (\Cref{fig:syntheticTraining}EF), are indicative of possibly ill-defined cases. 
Using elongation as a reverse proxy for geometric ambiguity confirmed that shapes with low elongation tended to yield less accurate and more variable predictions (\Cref{fig:sup-ambiguity16AE}). Given the variability of division orientations in the apical external domain \citep{Laruelle_2022}, it is also possible that the training set size was not sufficient to learn in a robust manner the relation between cell shape and plane positioning.
The heterogeneous distribution of accuracy for 16AI predictions (\Cref{fig:individualCells}A) was consistent with the heterogeneity of observed division patterns in this domain and with the inability of the centroid rule to reproduce some of them \citep{Laruelle_2022}.
Interestingly, the 16BI divisions were correctly predicted by the model (\Cref{fig:individualCells}AB), suggesting that, although the centroid rule does not account for these divisions \citep{Laruelle_2022}, they could be subtended by an alternative geometrical principle. In the four 16C domains, the predictions in native mother cells were consistent with predictions in reconstructed ones (\Cref{fig:individualCells}CD).

Altogether, our results show that the masked-UNet can link real cell shapes to observed division plane positioning when such a relation unambiguously exists. Interestingly, our results on 16BI division patterns also illustrate how a purely data-driven approach may yield diverging results from the alternative approach of testing an \textit{a priori} given division rule.

\subsection{A single model accounts for unequivocal division patterns  in \textit{A. thaliana} early embryo}

Having established the masked-UNet ability to learn well-defined embryo division patterns on a per stage and per domain basis, we evaluated its ability to learn a single rule covering all these patterns. The 1C, 16AI, and 16AE patterns, corresponding to ambiguous or near-ambiguous cases, were excluded from the training set.

\begin{figure}
\includegraphics[width=\textwidth]{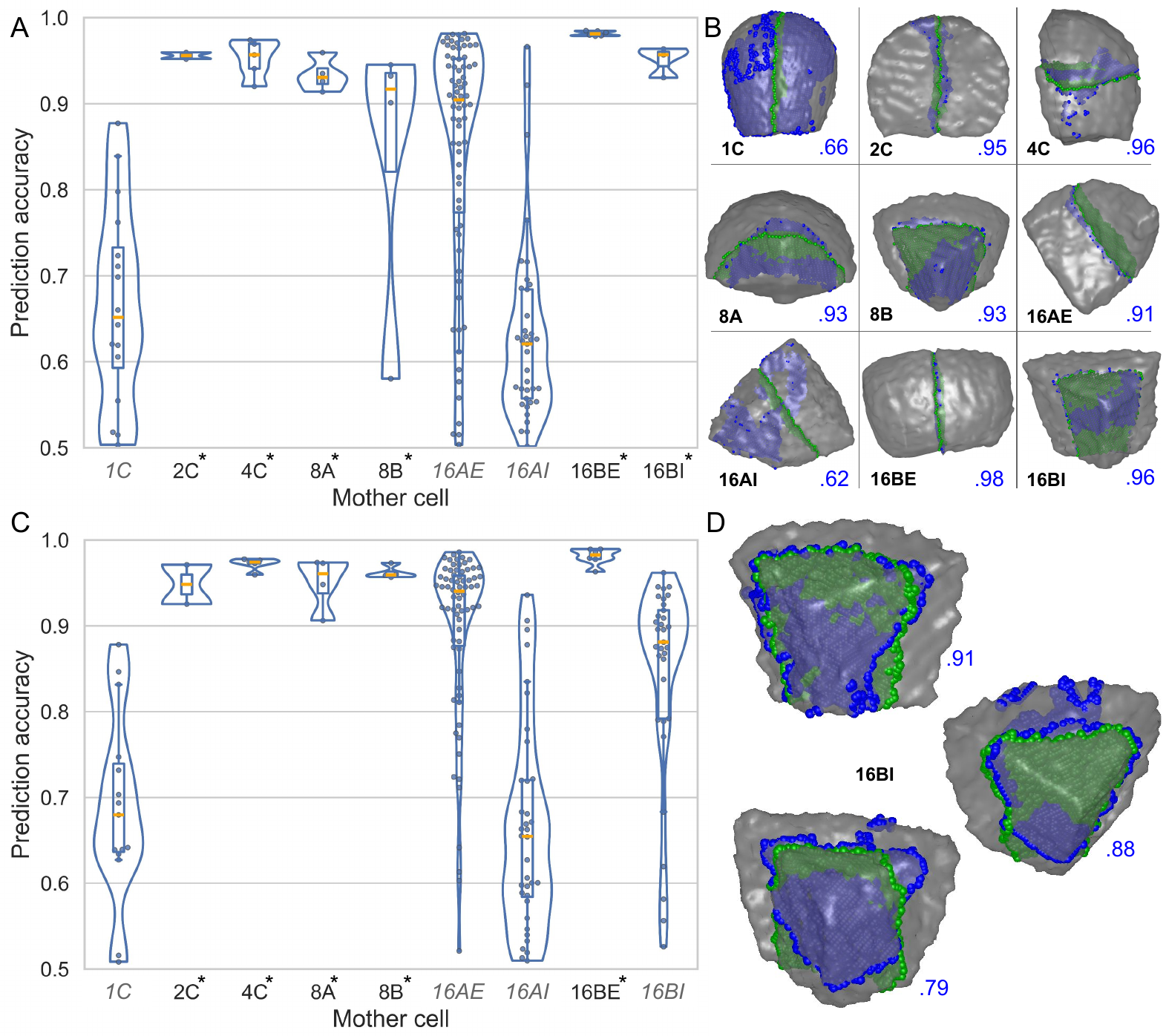}
\caption{
Learning stereotyped division patterns in \textit{Arabidopsis thaliana} embryo.
A single model was trained on stereotyped division patterns between 1C and 16C. 
(A)~Prediction accuracy for all generations and domains between 1C and 16C. Cases represented in the training set are marked in bold with an asterisk ($^*$).
(B)~Model predictions. The selected cases correspond to median accuracy. Numbers give the corresponding accuracies.
(C)~Prediction accuracy when removing the 16BI division patterns from the training set. Cell volumes were normalized (see Material and Methods).
(D)~Model predictions in 16BI cells. The three cases correspond to 1rd, 2nd, and 3rd quartiles of the accuracy distribution. \textit{Green}: target (observed) division plane; \textit{Blue:} predicted division plane.
}
\label{fig:EmbryoAll}
\end{figure}

On the test set, all well-defined division patterns were accurately predicted (\Cref{fig:EmbryoAll}AB). This shows the model can learn a geometrical division rule fitting all unequivocal division patterns, and suggests that a shared geometrical principle could underlie these patterns despite their diversity. Remarkably, the 16BI cells, which are not predicted by the centroid rule \citep{Laruelle_2022}, were accurately predicted by the trained network. Hence, the model could have captured a more general division rule than the centroid rule. Within the set of patterns excluded from training, the 1C and 16AI divisions were globally inaccurately predicted, in accordance with their pronounced ambiguities. Surprisingly, the nearly-ambiguous 16AE patterns were globally accurately predicted (\Cref{fig:EmbryoAll}AB) with comparable accuracy distribution to the 16AE specialized model (\Cref{fig:individualCells}A). This indicates these patterns share enough similarities with the well-defined ones to be captured by the model, in accordance with previous results showing the 16AE patterns also followed the centroid rule \citep{Laruelle_2022}.

To further corroborate that 16BI division patterns could be unified with other unambiguous patterns, we trained a model on all well-defined division patterns, excluding the 16BI cases. 
While the model accurately predicted the well-defined division patterns used for training (and most of 16AE cases), predictions in 16BI cells were incorrect (\Cref{fig:sup-LOO16BI_Default}). 
This was rather unexpected given the similarity between 16BI and 8B division patterns (unequal and periclinal divisions of triangular prism shapes). Given 16BI cells are on average the smallest ones over the non-equivocal considered cases (\Cref{fig:sup-generation-properties}A), we hypothesized size differences could hamper the generalization of model predictions.
We thus trained the model after resizing cell shapes to a common, fixed volume (see Material and Methods). In this condition, the accuracy of the model on 16BI patterns was significantly improved (\Cref{fig:EmbryoAll}CD), confirming the influence of cell size in model predictions and the importance of having consistent data between training and inference stages. The 16BI cell divisions have a unique distribution of volume-ratio compared with other cases (\Cref{fig:sup-generation-properties}B). The discrepancy in volume-ratio with the training set could explain the lower accuracy observed in test 16BI shapes, since volume-ratios in predictions were close to that of 8A and 8B patterns but not to the ones observed in 16BI (compare \Cref{fig:sup-VR_LOO16BINorm} to \Cref{fig:sup-generation-properties}B). Contrary to the size effect, there is unfortunately no possibility to normalize for this effect. Altogether, in spite of this inherent difficulty, our results strongly suggest a common geometrical principle could underlie the non-equivocal division patterns up to 16C in \textit{A. thaliana} embryo. 

\section{Discussion}

In this paper, we proposed a novel approach using deep learning to establish a relation between mother cell geometry and division plane position and orientation. Investigating such relations is crucial in developmental studies, in particular in plants where division is a key determinant of tissue organization. Geometrical division rules are useful because they point to invariant principles that can unify division patterns across different species, tissues, developmental contexts, and cell shapes \citep{Couturier2025}. In addition, they can reflect shared physical principles. Historically, a few geometrical division rules have been formulated from experimental observations. Some rules are more general than others. For example for equal divisions, the Sachs rule \citep{Sachs_1887}, stipulating that new division walls intersect at 90 degrees previous walls, is implied by Errera's rule, the plane area minimization principle. Consequently, failure to capture observed division patterns by a given rule does not allow to reject the existence of a geometrical relation between cell shape and plane positioning. The approach we introduced here addresses this issue. We show that the existence of a geometrical rule can be tested through the ability of a deep neural network to learn and generalize division plane positioning from cell geometry.

Inspired by our previous work on modeling cell division patterns in an image-based framework \citep{Moukhtar_2019,Laruelle_2022}, we cast division prediction as a cell partitioning problem and relied on a UNet to learn this task. Though UNet is a state-of-the-art CNN architecture for the related problem of image segmentation, we show several modifications were needed to reach invariance or quasi-invariance regarding daughter cell labels, cell position and cell orientation. The main advantages of an image-based formalism are the possibility to train and evaluate powerful CNNs from experimentally observed patterns and the preservation of geometrical complexity in cell shapes and in division patterns. The major counterpart of the UNet-based architecture are the deterministic nature of its predictions and the training procedure with a single division pattern associated to each input cell. Accordingly, the network was unable to learn division rules on cell shapes with multiple equivalent divisions. One way to address this limitation when learning a known rule would be evaluating the adequacy to the rule rather than the similarity to a reference pattern, or by associating multiple reference patterns to a given cell shape. The deterministic nature of the predictions also implies that the current architecture should not be able to capture stochastic division rules, where divisions follow a geometry-dependent probability distribution as shown in some plant tissues \citep{Besson_2011}. This could be addressed by turning to stochastic networks such as generative adversarial networks \citep{Goodfellow2014} or diffusion models \citep{SohlDickstein2015,Ho2020}. Both can learn with conditioning, which is required to specify cell geometry.

Evaluating the approach on \textit{A. thaliana} embryo cells revealed that the model could learn various division patterns over a large range of real cell shapes in spite of their morphological variability.
Except for divisions at 1C and in the internal apical domain at 16C, which corresponded to ill-defined cases impossible to learn for the network,  the masked-UNet successfully reproduced all well-defined division patterns observed up to 16C stage. It had been shown previously that many division patterns up to this stage could be recapitulated by the centroid rule---the minimization of division plane area conditioned on passing through the centroid \citep{Moukhtar_2019,Laruelle_2022}. However, this rule had failed to predict division plane positioning in the internal basal domain at 16C \citep{Laruelle_2022}. The present study allows to revisit these results since the masked-UNet unifies the 16BI division with other well-defined patterns at the same or previous stages, suggesting that a common geometrical principle may subtend all these patterns. A further implication is that the centroid rule could be a specific formulation of a more general rule.
Unfortunately, the masked-UNet, with a few million parameters, is an opaque model from which extracting human-interpretable rules remains challenging. Future work might involve considering more interpretable architectures, enabling deriving explicit rules and formulating new hypotheses.

Deep learning has been used in multiple domains to solve complex tasks with higher accuracy than feature-based methods. In biological imaging, most applications have dealt with image restoration, segmentation, analysis, or classification \citep{Hallou_2021}. In our approach, we train deep networks to reproduce patterns that are given by observations, to evaluate possible links between cell geometry and division plane positioning. By doing so, we shift the focus from the prediction itself to the ability to predict. Works as the present one that contribute to widen the potential of deep learning for new insights in developmental and cell biology are only emerging. For example, \citet{Li_2024} trained diffusion models to learn and characterize geometrical relationships between stress fibers and cell shape. Beyond differences in network architectures, the ability to exploit 3D image data and to predict segmented patterns rather than localization probabilities are the main distinctive features between our work and that study.

While the full implications of our approach remain to be explored, we believe it offers a valuable perspective in the study of cell division from a geometrical standpoint.
For instance, it can help determine whether geometric information alone is sufficient for predicting division behavior, or if additional cues have to be considered.
Because our 3D discrete formalism is congruent with imaging data, it could easily integrate additional spatial information involved in controlling division orientation such as polarity cues, chemical gradients, cell topology \citep{Carter2017}, or mechanical stresses \citep{Louveaux_2016}.
Hence, this approach offers a promising foundation for developing a general and versatile model of cell division, which could serve as a building block for \textit{in silico} simulations of tissue morphogenesis. Ultimately, such an approach could become part of the computational synthetic biology toolkit \citep{Kim_2020} and prove instrumental in engineering self-developing biological systems—much like how evolutionary algorithms were employed to predict the behavior of \textit{in vitro} organoids in \citep{Kriegman_2020}.

\section{Acknowledgments}

This work has benefited from a French State grant (Saclay Plant Sciences, reference n° ANR-17-EUR-0007, EUR SPS-GSR) managed by the French National Research Agency under an Investments for the Future program integrated into France 2030 (reference n° ANR-11-IDEX-0003-02) (funding to AD).
This project was provided with computer and storage resources by GENCI at
IDRIS thanks to the grant 2024-AD011014549 on the supercomputer
Jean Zay's V100 partition. We thank Emmanuel Faure (LIRMM, Montpellier, France) for feedback during the early stages of this work and for comments on the manuscript. We also thank Jasmine Burguet (IJPB, Versailles, France) for her comments on a previous version of the manuscript.
\section{Material and methods}

\subsection{Synthetic cells}

Two categories of synthetic cell shapes, cuboids and ellipsoids, were generated as binary masks in 3D images.
Cuboids are an idealization of \textit{A. thaliana} embryo cells, which statistically tend to have six topological cuts \citep{Laruelle_2022}. Cuboids and ellipsoids have distinct, complementary features.
Cuboid surfaces are flat almost everywhere, with curvature singularities at edges. On the opposite, ellipsoid surfaces have defined, non-zero curvature everywhere.
Both cubois and ellipsoids had their principal axes parallel to the image coordinate axes.
Cuboids were defined by the half-length $a$ of their longest axis, the half-length $b$ of their medium axis, and the half-length $c$ of their shortest axis. 
Ellipsoids were defined by their equation $\frac{x^2}{a^2} + \frac{y^2}{b^2} + \frac{z^2}{c^2} = 1$, where $a, b$ and $c$ are the half-lengths of the longest, intermediate, and shortest axes. 

The datasets were constructed by sampling shape elongation $e$, flatness $f$, and volume $v$ from the following uniform distributions: $e \sim \cal{U}$(1, 3), $f \sim \cal{U}$(1, 3), and $v \sim \cal{U}$(24000, 120000 voxels).
Parameters $a,b$ and $c$ were then obtained as: 
$$
    \left \{
   \begin{array}{l c l}
      a = (fve^{2})^{1/3}\\
      b = \frac{a}{e}\\
      c = \frac{b}{f}
   \end{array}
   \right .
$$
for cuboids and
$$
    \left \{
   \begin{array}{l c l}
      a = (\frac{3fve^{2}}{4\pi})^{1/3}\\
      b = \frac{a}{e}\\
      c = \frac{a}{ef}
   \end{array}
   \right .
$$
for ellipsoids.

Division according to Errera's rule in ellipsoids and cuboids was obtained by positioning the division plane through the cell centroid  orthogonally to the principal axis. In these specific shapes, this was indeed congruent with plane area minimization under symmetric division. Division according to the ``Anti-Hertwig'' division rule was generated by positioning the division plane through the shape centroid orthogonally to the smallest axis. Each synthetic dataset contained 500 pairs of mother/daughter shapes. The distribution between training, validation, and test sets was 400/50/50.

\subsection{Embryo cells}

Images of fixed \textit{A. thaliana} embryos at different stages of their development were acquired using 3D confocal imaging following labeling of cellular walls. Detailed experimental and image segmentation protocols have been described earlier \citep{Moukhtar_2019,Laruelle_2022}. Resampling was applied to reach an isotropic voxel size of 0.35~µm along each axis. Because of the fixation step in the acquisition protocol, it was not possible to visualize the temporal evolution of embryos. Pairs of mother/daughter cells were thus obtained by identifying sister cells and merging them back into their mother cells \citep{Moukhtar_2019}. A corollary of this procedure is that the mother cell volume and shape are identical before and after cell division.

Each pair of mother/daughter cells was annotated based on the generation and spatial domain (A: apical, B: basal, I: internal, E: external) of the mother cell in the embryo, ranging from 1C to 16C (1C, 2C, 4C, 8A, 8B, 16AE, 16AI, 16BE, 16BI). The dataset contained a total of 375 division samples (\Cref{tab:embryo-cells}).

Divisions following Errera's rule (symmetric division with area minimization of the division plane) were simulated in embryo cells using a stochastic partitioning algorithm previously described \citep{Moukhtar_2019,Laruelle_2022}. Initially, each voxel of the mother cell mask was randomly assigned the label of one of the  daughter cells. This random configuration was then iteratively modified using a Metropolis algorithm to stochastically minimize the area of contact between the two daughter cells. Details, code and executable of division simulation can be found in \citet{Laruelle_2022}.

\subsection{Masked-UNet architecture}

The UNet model architecture was chosen to have a depth of 5 levels, corresponding to a receptive field at the lowest layer of 140 voxels, meaning that neuron activations at this level are potentially influenced by the input voxels inside a window of 140 voxels \citep{Araujo_2019}. This receptive field was selected to be larger than the largest cell of our dataset, ensuring that the lowest model layers could integrate the information of all the voxels located in the mother cell. 

In the masked-UNet, the binary mask of the mother cell is propagated at all levels and channels of the network (\Cref{fig:initialModifications}C) and used to mask convolution operations (\Cref{fig:sup-propagatedMask}B).
To maintain the mask and the activation tensor to the same shape across the different levels, the maxPooling operation was also performed on the mask. However, no upsampling was applied to the mask to avoid altering the cell shape. Instead, the mask was propagated between the encoding and decoding parts of the network through the skip-layer connections (\Cref{fig:initialModifications}C). Additionally, the group normalization operation was modified to ignore the masked activations.

A common practice to break the inherent hierarchy between labels is to encode them through channels. We used two channels to encode the mother cell input, to represent the labels 0 and 1 in a "one-hot" fashion, and two channels for the division output, one for each daughter label 1 and 2. 
Doing so only predicts the partitioning of the two daughter cells. The background voxels of the prediction were then set to 0 after model inference by using the binary input mask, thus restricting the prediction to the space of the mother cell.

\subsection{Training the masked-UNet}

In all experiments, the datasets were split into 3 subsets with different purposes: training ($\sim$80\%), validation ($\sim$10\%), and testing ($\sim$10\%). The training set was used during the training phase to tune the network's parameters. The validation set was used during the training phase to ensure that the network was not overfitting the training set. Lastly, the test set was used to assess the global performance of the trained network on data not seen during the training phase.

During the training phase, we used batches of size 16.
The network's parameters were updated using back-propagation with the Adam optimizer \citep{Kingma_2017} and a starting learning rate of $10^{-4}$. 
A modified raw cross-entropy (CE) loss function was used to obtain a quantitative measure of the prediction  error restricted to the mother cell mask:
\begin{equation}
    \mathrm{CE}(\hat{y}, y) = - \frac{1}{\sum_{n=1}^{N}H(y_{n})}\sum_{n=1}^{N}  \log\frac{\exp(\hat{y}_{n,y_n})}{\sum_{k=1}^{2}\exp(\hat{y}_{n,k})} \times H(y_{n})
\label{eq:cross-entropy}
\end{equation}
where $y_n$ is the expected label for voxel $n$, $\hat{y}_{n,k}$ is the model prediction that voxel $n$ be of label $k$, and $H$ is the Heaviside function: $H(y_n)=1$ for $y_n>0$, otherwise $H(y_n)=0$.
Similarly, the raw accuracy (ACC)  was defined as the proportion of correctly labeled voxels in the mother cell mask:
\begin{equation}
    \mathrm{ACC}(\hat{y}, y) = \frac{ \sum_{n=1}^{N} \mathds{1}_{\arg\max_{k} (\hat{y}_{n,k}) = y_n}\times H(y_{n})}{\sum_{n=1}^{N}H(y_{n})}
\label{eq:accuracy}
\end{equation}
where $\mathds{1}_{\cdot}$ is the indicator function of the event where the expected label $y_n$ equals the predicted label $\arg\max_{k}\hat{y}_{n,k}$. Finally, to enforce insensitivity to the labeling of daughter cells, label-symmetric versions of the loss and accuracy metrics were introduced to select the best values computed with the target pattern $y$ and its symmetric version $\bar{y}$ obtained by swapping the daughter labels (i.e., $\bar{y}=y/2(7-3y)$):
\begin{eqnarray}
\mathrm{Loss}(\hat{y},y) &=& \min\left\{\mathrm{CE}(\hat{y},y),\mathrm{CE}(\hat{y},\bar{y})\right\}\\
\mathrm{Accuracy}(\hat{y},y) &=& \max\left\{\mathrm{ACC}(\hat{y},y),\mathrm{ACC}(\hat{y},\bar{y})\right\}
\end{eqnarray}

The training was run for several epochs (one epoch corresponds to a full observation of the training set) until no longer noticeably reducing the loss value.

The constitution of a batch requires all images to be of the same shape (number of rows, columns, and planes). A common approach consists of resizing images accordingly. However, this  modifies the spatial scale of the images and may deform the input cell shapes. Therefore, having determined spatial resolution to be an important feature of our model, we decided instead to  pad the border of the smaller images to reach the largest image size in each batch. This method required more memory space during training but preserved the image spatial calibration and cell shapes.

Training on individual embryo domains (\Cref{sec:individual-training}) was performed using a 10-fold approach. This method runs 10 training folds, each of which is  evaluated on a different part (10\%) of the dataset such that the whole dataset is tested during the whole process. This 10-fold approach resulted in a total of 90 individual training (10 folds, 9 generations or embryo domains considered), each conducted for 15000 training steps.

Model inferences in embryo native (non-reconstructed) mother cells were computed after size normalization. The binary images of input cells were scaled isotropically with nearest neighbour interpolation. A scaling factor was computed for each embryo generation and domain, to ensure that the scaled native cells had on average the same volume (in voxels) as the reconstructed cells used during training.

Similarly, the 16BI leave-one-out experiment with standardized cell volumes consisted in rescaling each cell (in both training and test sets) to reach as closely as possible a target volume $V_t$, using an isotropic scaling factor $\sqrt[3]{\frac{V_t}{V}}$, where $V$ was the original volume of the cell. To minimize the average amplitude of scaling applied to each cell, $V_t$ was set to the initial average volume of all cells used during training (23320 voxels).

\bibliographystyle{biology}
\bibliography{library}

\appendix
\setcounter{figure}{0} 
\setcounter{table}{0} \renewcommand\thefigure{S\arabic{figure}}
\renewcommand\thetable{S\arabic{table}}
\cleardoublepage

\section*{Supplementary Tables and Figures}

\cleardoublepage

\begin{table}
    \centering
    \begin{tabular}{c|c|c|c|c|c|c|c|c|c}
       Domain & 1C & 2C & 4C & 8A & 8B & 16AE & 16AI & 16BE & 16BI \\\hline
       Divisions &  16 & 22 & 48 & 40 & 40 & 72 & 37 & 65 & 33\\
       Mean \#voxels & 60386 & 36069 & 24842 & 19350 & 22449 & 19122 & 12505 & 25917 & 15401\\
       Std-dev. \#voxels & 16260 & 4765 & 6027 & 4806 & 5856 & 5479 & 5108 & 7803 & 4764 \\
    \end{tabular}
    \caption{Extracted embryo cell divisions, referenced by their mother cell generation and localization within the embryo, and their number of occurrences. Average and standard deviation of the number of voxels in mother cells are also given.}
    \label{tab:embryo-cells}
\end{table}

\cleardoublepage

\begin{figure}[p]
\includegraphics[width=\linewidth]{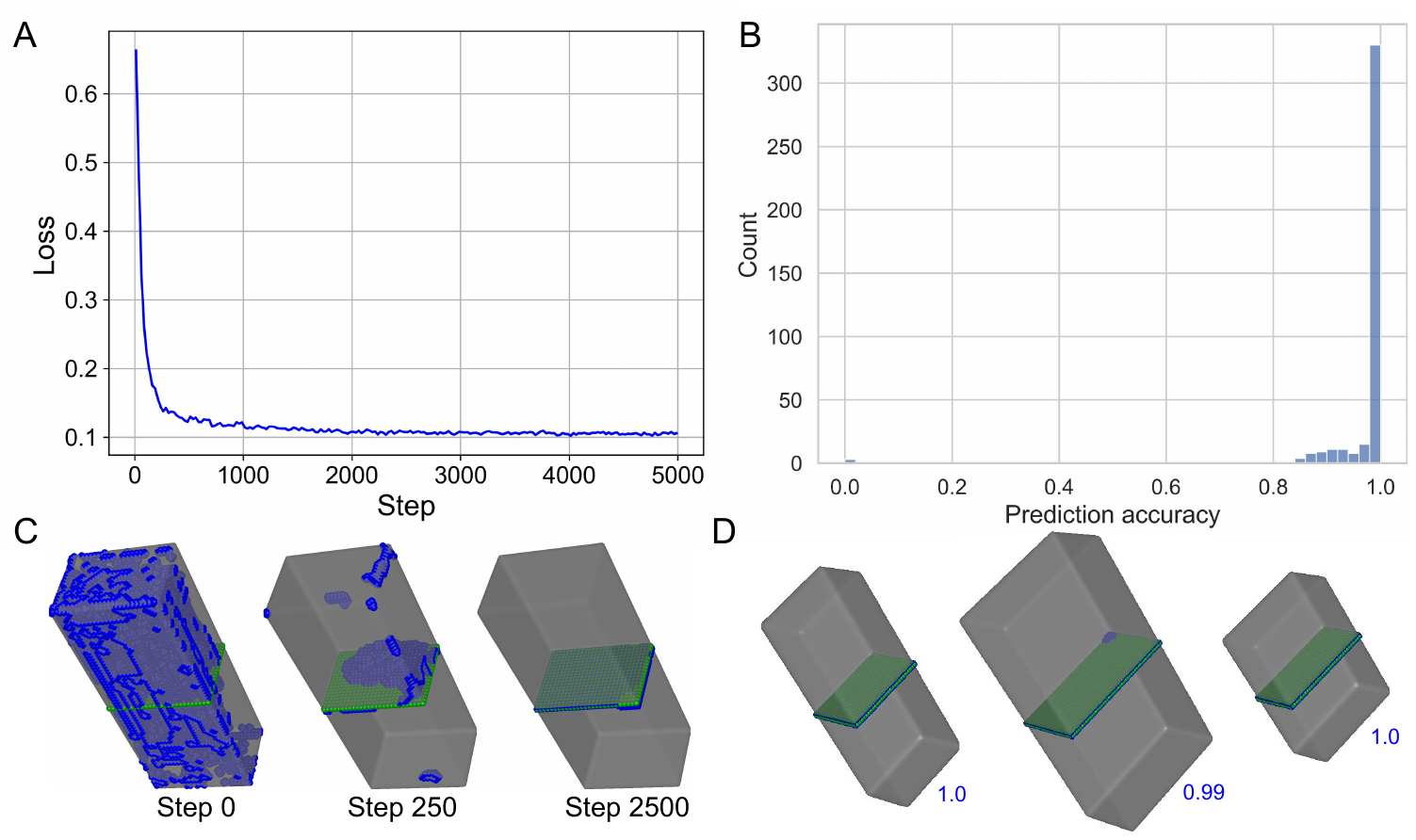}
\caption{Training the native UNet-3D to learn Errera's division rule in cuboids.
(A)~Training loss.
(B)~Distribution of prediction accuracy over the training set (N=400).
(C)~Evolution of the predicted plane in a given shape during training. \textit{Blue}: Predicted division plane. \textit{Green}: target division plane.
(D)~Predictions in different shapes. Numbers: prediction accuracy.
}
\label{fig:sup-initialTraining}
\end{figure}

\cleardoublepage

\begin{figure}
\includegraphics[width=\linewidth]{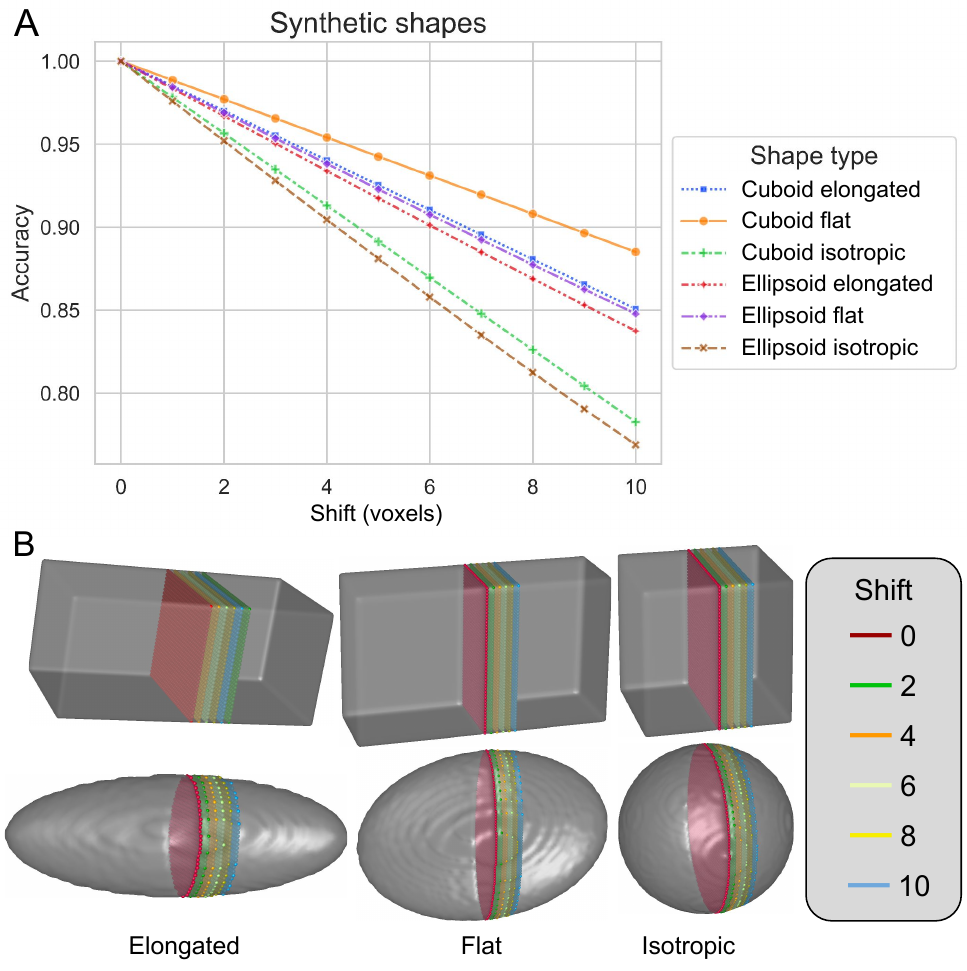}
\caption{Sensitivity of the accuracy metrics to controlled errors on the division patterns in synthetic shapes. Three shapes were considered for cuboids and for ellipsoids. In each shape, a symmetric division according to Errera's rule (\textit{Red}) was taken as reference. The reference plane was systematically shifted by applying an erosion to one of the daughter cells. The accuracy of the resulting patterns was computed with regards to the reference pattern.
(A)~Evolution of accuracy as a function of plane shift (in voxels).
(B)~Examples of division planes obtained for different  plane shifts.}
\label{fig:sup-accuracy-erosion-synthetic}
\end{figure}

\cleardoublepage

\begin{figure}[t]
\includegraphics[width=\linewidth]{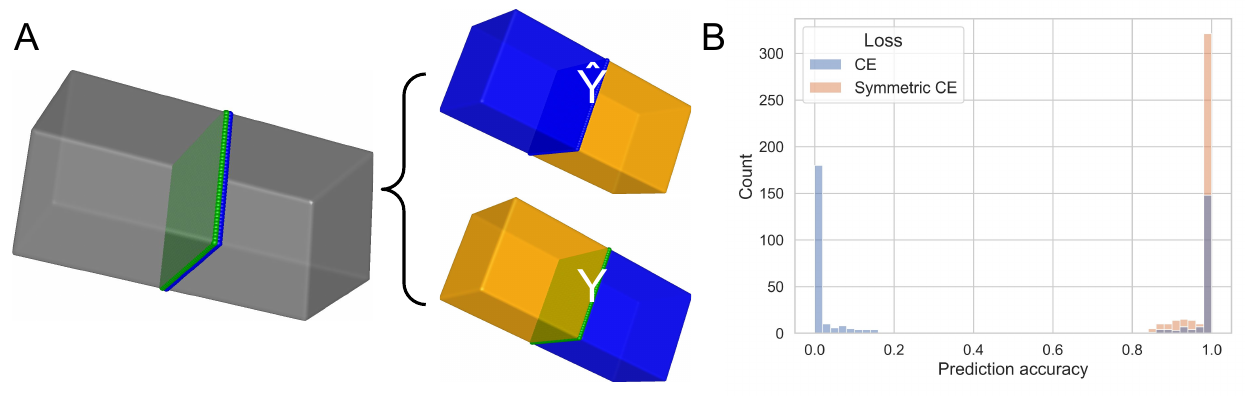}
\caption{Enforcing invariance to daughter cell labeling. (A)~The same division plane (\textit{Left}) can be represented by two different patterns with swapped daughter labels (\textit{Right}). However, without symmetrization, a correct prediction such as $\hat{Y}$ would be assigned the lowest possible scores (loss and accuracy) because of label mismatch to the target $Y$. (B)~Without symmetrization of loss (CE: cross-entropy) and accuracy (\textit{Blue}), roughly half of the predictions are incorrect. Symmetrization of loss and accuracy solves this issue (\textit{Orange}).}
\label{fig:sup-label-invariance}
\end{figure}

\cleardoublepage

\begin{figure}[t]
\includegraphics[width=\linewidth]{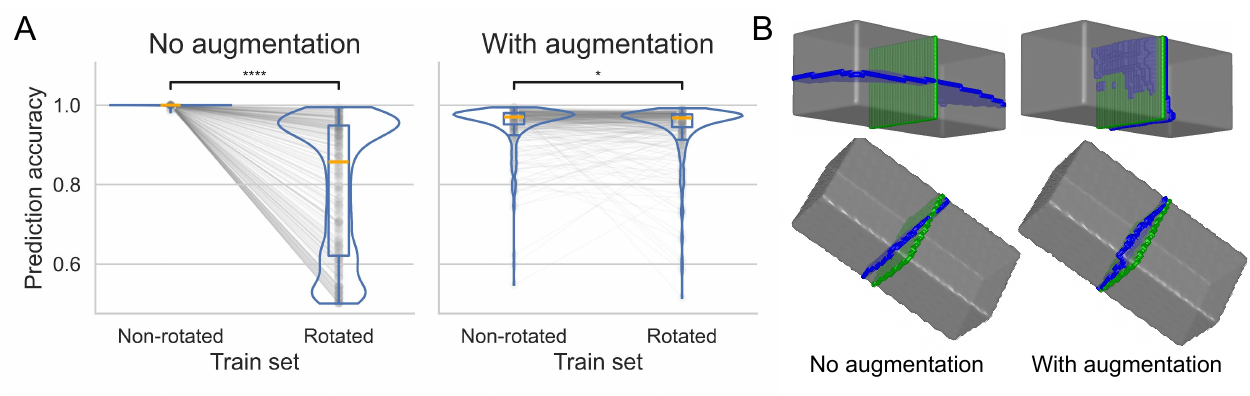}
\caption{Enforcing rotation equivariance using data augmentation.
(A)~Effect of random rotations of test shapes on accuracy for models trained without (\textit{Left}) and with (\textit{Right}) rotation augmentation. Paired Wilcoxon test: $P$=2.73e-67 (\textit{Left}) and 0.026 (\textit{Right}) (N=400). 
(B)~Predictions in a same shape at two different orientations, with models trained without (\textit{Left}) and with (\textit{Right}) rotation augmentation. \textit{Blue}: target plane; \textit{Green}: predicted plane.
}
\label{fig:sup-rotation-invariance}
\end{figure}

\cleardoublepage

\begin{figure}
\centering\includegraphics[width=0.8\linewidth]{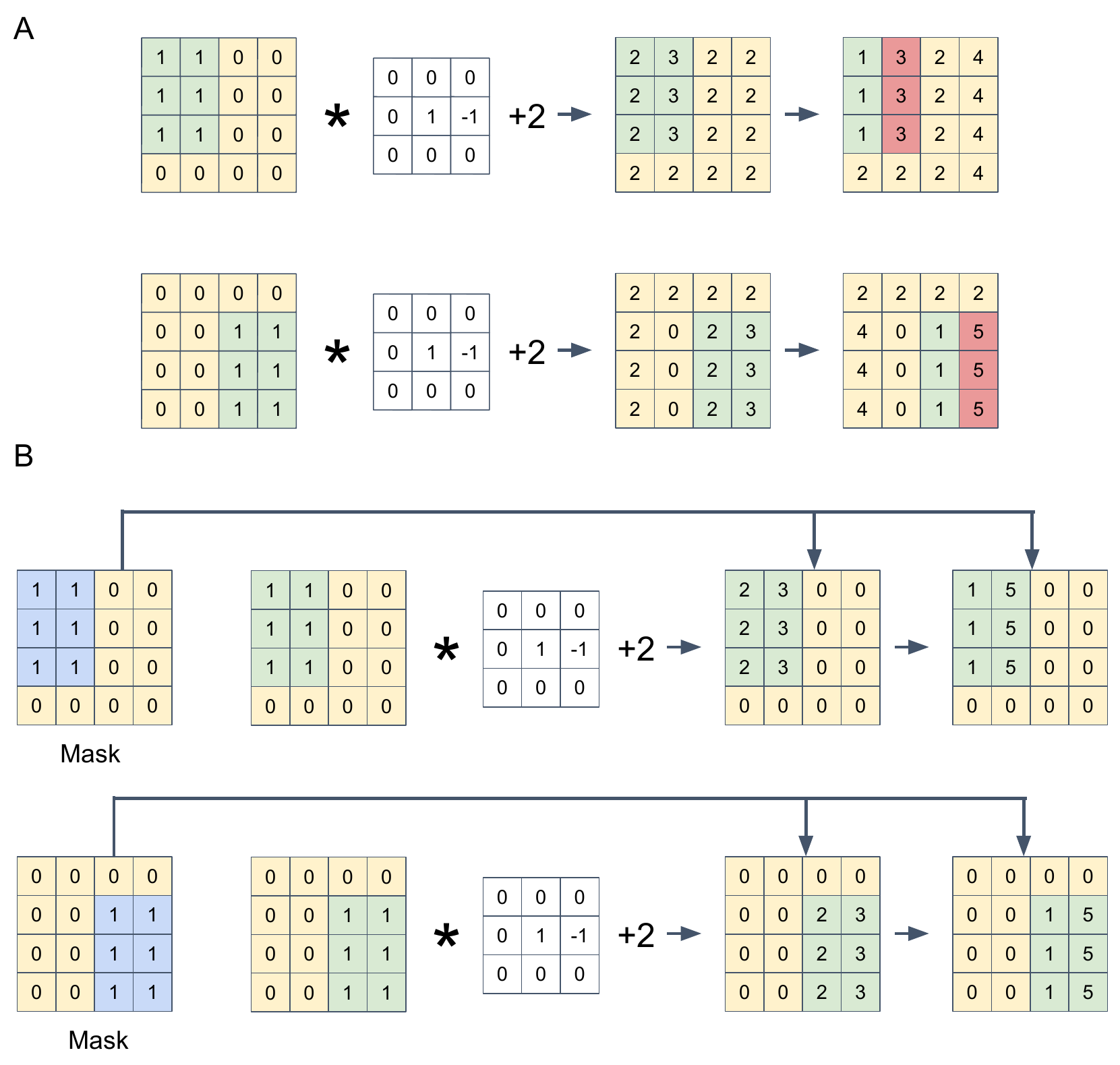}
\caption{Border artefacts in convolutional neural networks and their correction by mask propagation.
(A)~A standard convolution iterated twice yields different activation maps depending on object position in the image. In this example, the convolution is a simple horizontal difference, with a bias of +2. The input object is defined by 1s over a background of 0s and is located either in top-left (\textit{Top row}) or bottom-right position (\textit{Bottom row}). Pixels located outside the image (not represented) are assigned null values.  Object values affected by position are highlight in red.
(B)~Masking the convolution outputs with a mask of the object removes sensitivity to position. After each operation, values located outside the mask are set to 0.}
\label{fig:sup-propagatedMask}
\end{figure}

\cleardoublepage

\begin{figure}
\includegraphics[width=\linewidth]{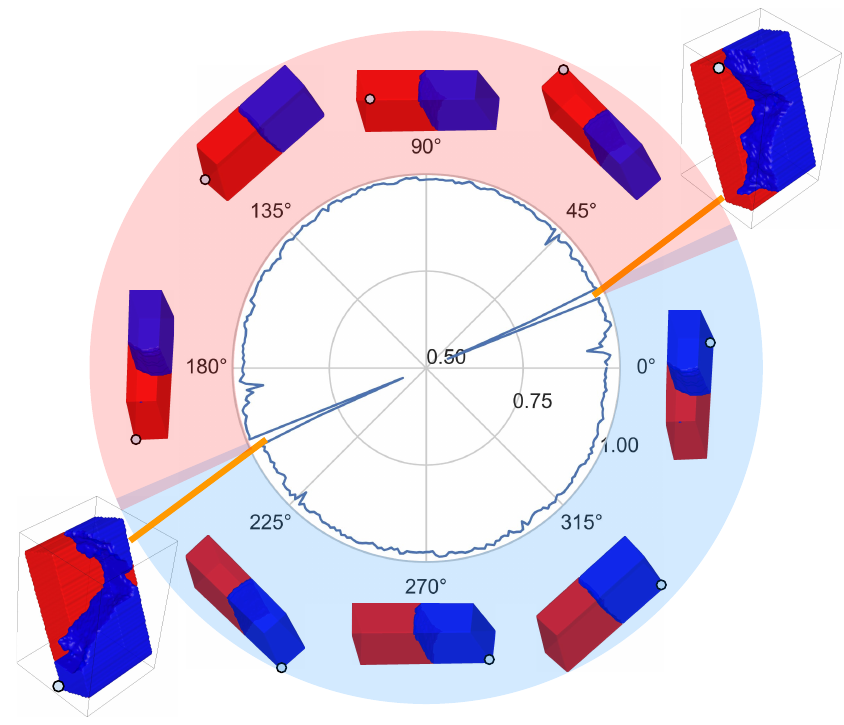}
\caption{Predictions in a given cell at different rotation angles along the X-axis reveals a label-ordering shifting point associated with lower accuracy. Prediction accuracy is plotted along the radial axis. The white dot indicates a specific corner of the cuboid mother cell. The label ordering phases are highlighted by the red and blue semi-circles, indicating the label of the white marker.}
\label{fig:sup-rotation-ambiguity}
\end{figure}

\cleardoublepage

\begin{figure}
\includegraphics[width=\textwidth]{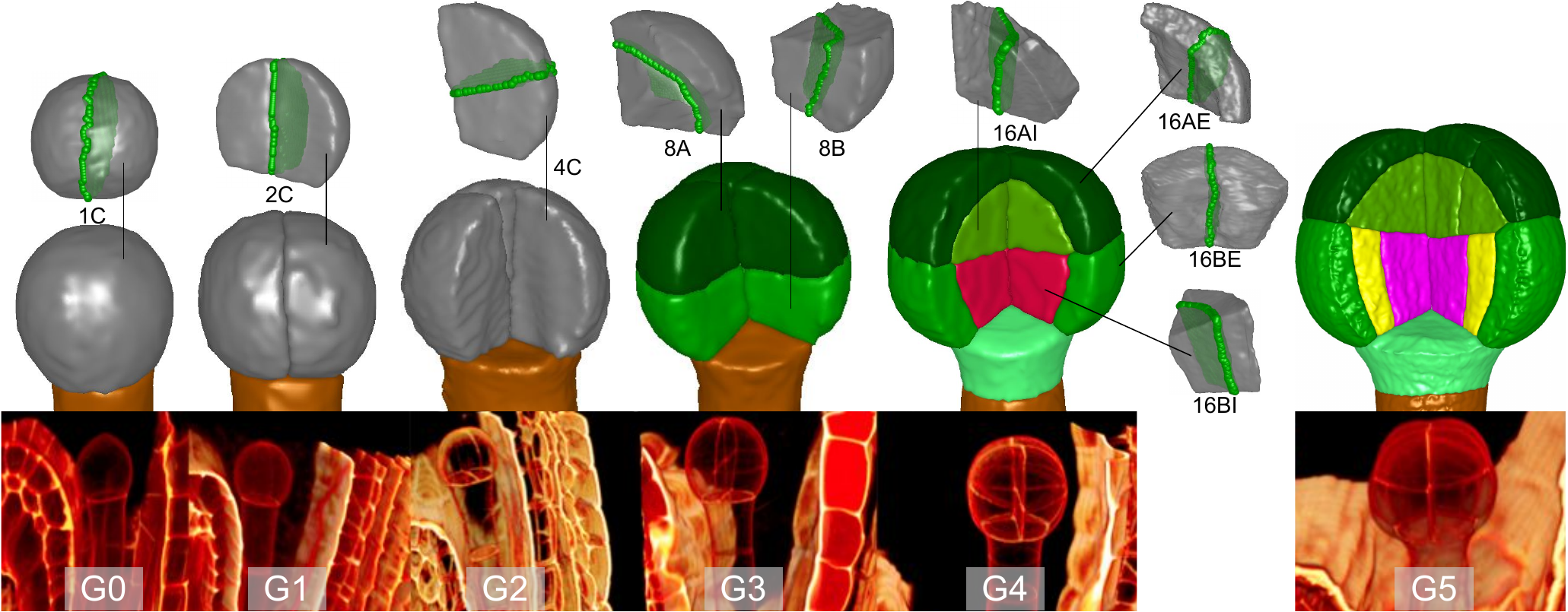}
    \caption{\textit{Arabidopsis thaliana} early embryo development.  \textit{Bottom row:} 3D confocal images of  embryos at different generations from G0 (1 cell) to G5 (32 cells). At the end of generation $n$, the embryo proper (not considering the suspensor) contains $2^n$ cells. \textit{Middle row:} 3D image segmentations at the different stages. \textit{Top row:} Individual cells from different stages and domains with their observed division planes. Cells are indexed by the number of cells at their generation (from 1 to 16) and their domain within the embryo (A: Apical, B: Basal, E: Extern, I: Intern). \textit{Green:} Observed division plane.}
    \label{fig:sup-embryo-development}
\end{figure}

\cleardoublepage

\begin{figure}
\includegraphics[width=\linewidth]{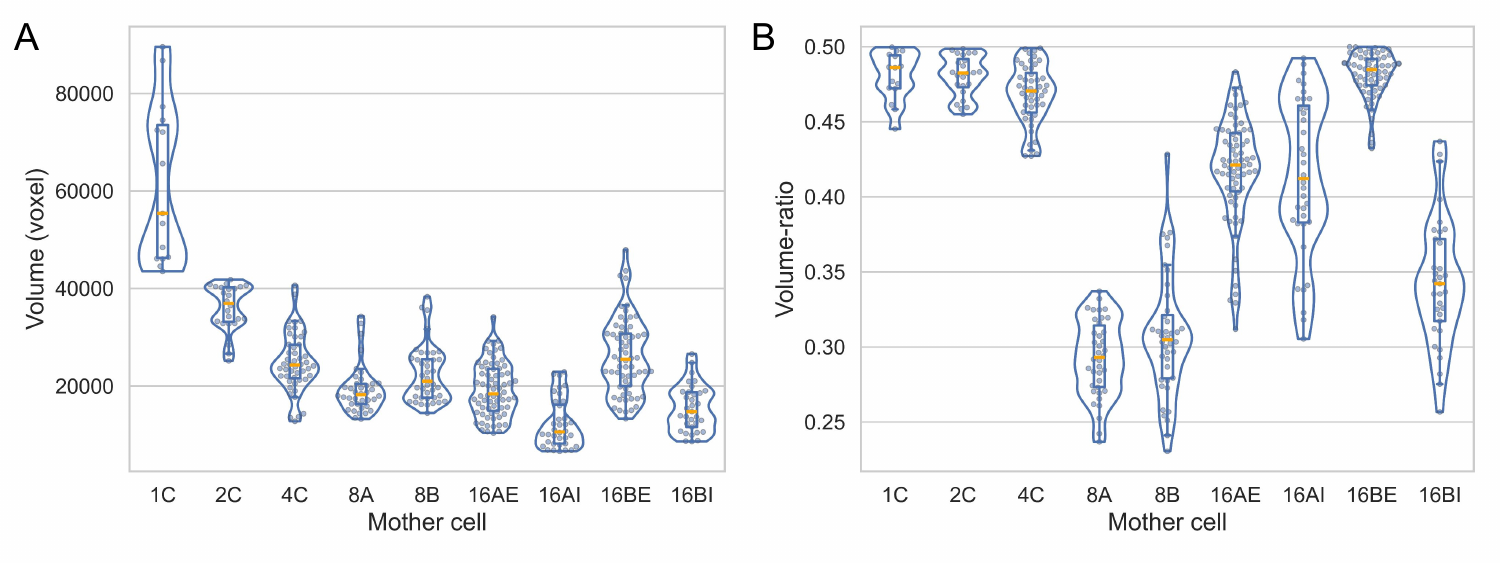}
\caption{Quantitative features of mother cells and divisions at different generations and in different embryo domains during early embryogenesis in \textit{Arabidopsis thaliana}. (A)~Mother cell volume (in voxels).
(B)~Volume-ratio of divisions (volume of the smallest daughter cell over the summed volumes of the two daughters).}
\label{fig:sup-generation-properties}
\end{figure}

\cleardoublepage

\begin{figure}
\centering
\includegraphics[width=\linewidth]{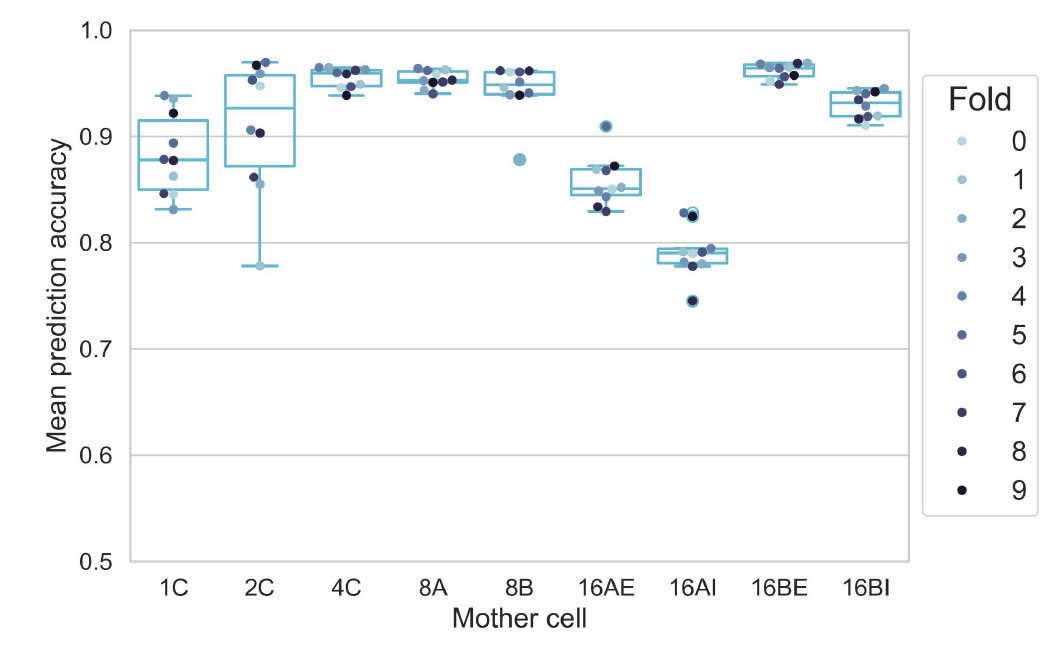}
\caption{Learning individual observed division patterns in \textit{Arabidopsis thaliana} embryo: per-fold results in a $k$-fold approach. For each generation and embryo domain between 1C and 16C stages, 10 distinct models were trained and tested with a different, non-overlapping partitioning of the dataset. Each point gives the average accuracy on the test set of each trained model.}
\label{fig:sup-kfoldMeans}
\end{figure}

\cleardoublepage

\begin{figure}
\includegraphics[width=\linewidth]{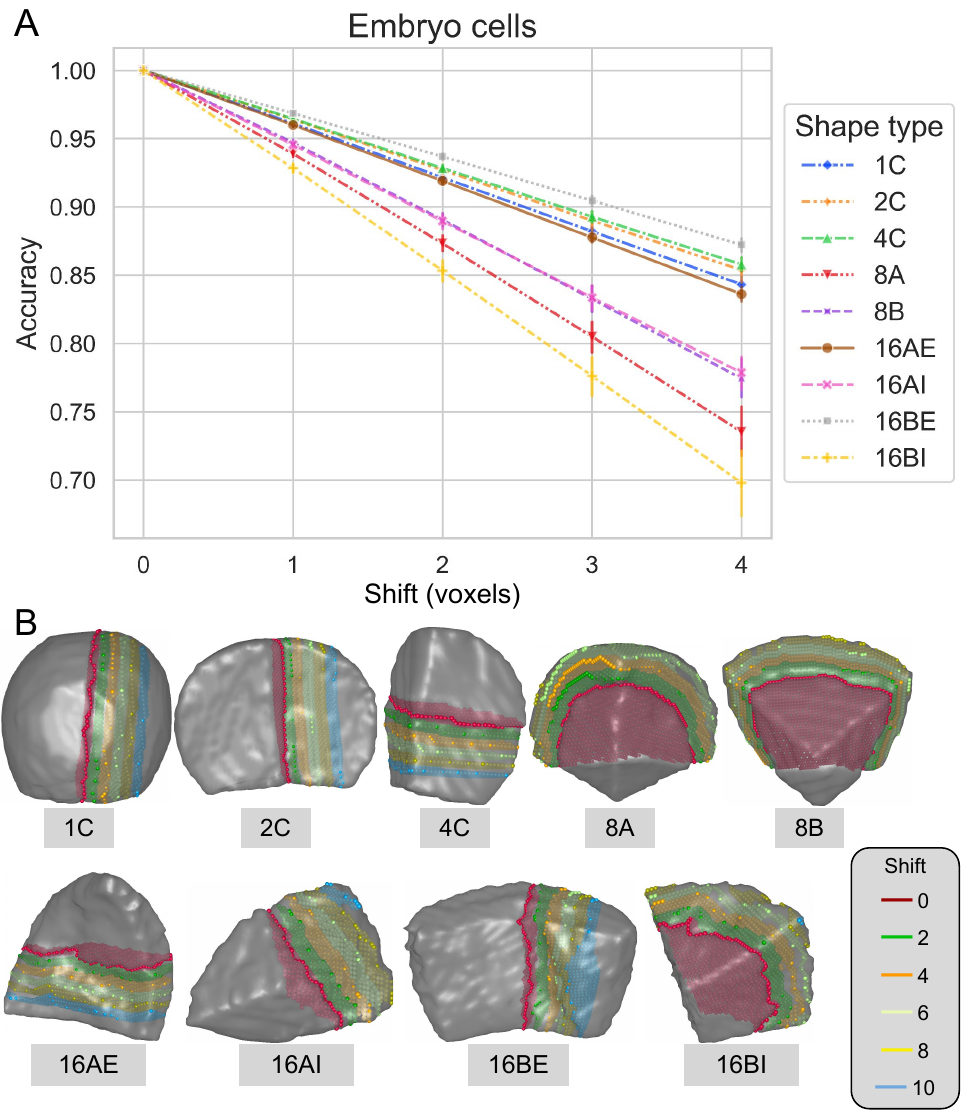}
\caption{Sensitivity of the accuracy metrics to controlled errors on observed division patterns in \textit{Arabidopsis thaliana} early embryo. Three divisions were considered at each stage and embryo domain between generations 1C and 16C. In each case, the observed division (\textit{Red}) was taken as the reference. The reference plane was systematically shifted by applying an erosion to one of the daughter cells. The accuracy of the resulting patterns was computed with regards to the reference pattern.
(A)~Evolution of accuracy as a function of plane shift (in voxels).
(B)~Examples of division planes obtained for different  plane shifts.}
\label{fig:sup-accuracy-erosion-embryo}
\end{figure}

\cleardoublepage

\begin{figure}
\includegraphics[width=\linewidth]{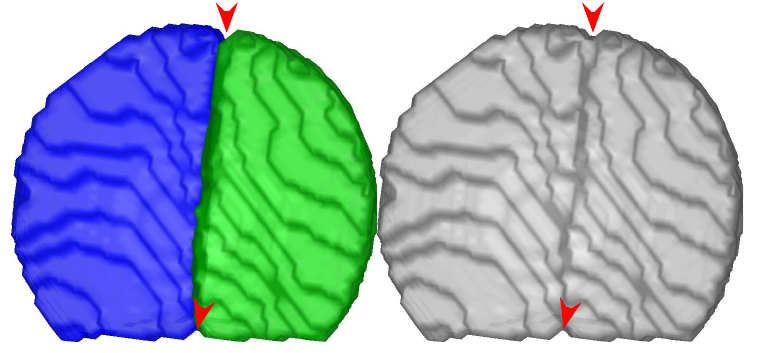}
\caption{Observed merging scar in daughter segmentation (\textit{left}) \textit{vs.} in reconstructed mother cell(\textit{right})}
\label{fig:sup-mergingScar}
\end{figure}

\cleardoublepage

\begin{figure}
\includegraphics[width=\linewidth]{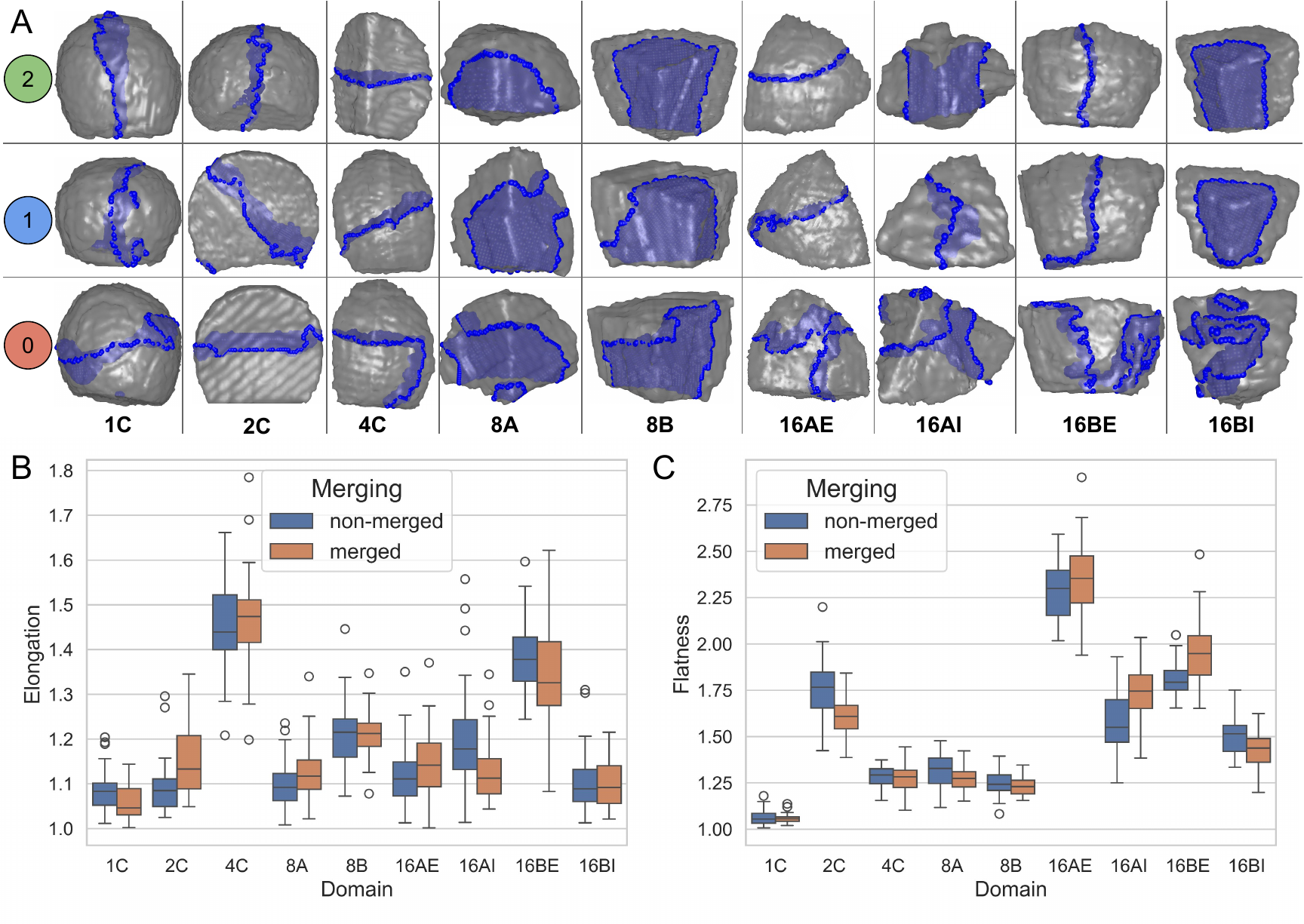}
\caption{Learning individual observed division patterns in \textit{Arabidopsis thaliana} embryo: validation on native (non-merged) cells.
(A)~Predicted division planes (\textit{Blue}) in native cells. Each column corresponds to a different embryo generation and spatial domain. In each case, the best model (over 10-folds) trained on reconstructed cells for the corresponding generation/domain was used. Each row corresponds to a visual score: \textit{0}: incorrect prediction; \textit{1}: noisy or partially correct prediction; \textit{2}: correct prediction. 
(B)~Cell shape elongation in native (non-merged) and reconstructed (merged) cells.
(C)~Cell shape flatness in native (non-merged) and reconstructed (merged) cells.
}
\label{fig:sup-noMerging}
\end{figure}

\cleardoublepage

\begin{figure}
\centering   
\includegraphics[width=.618\linewidth]{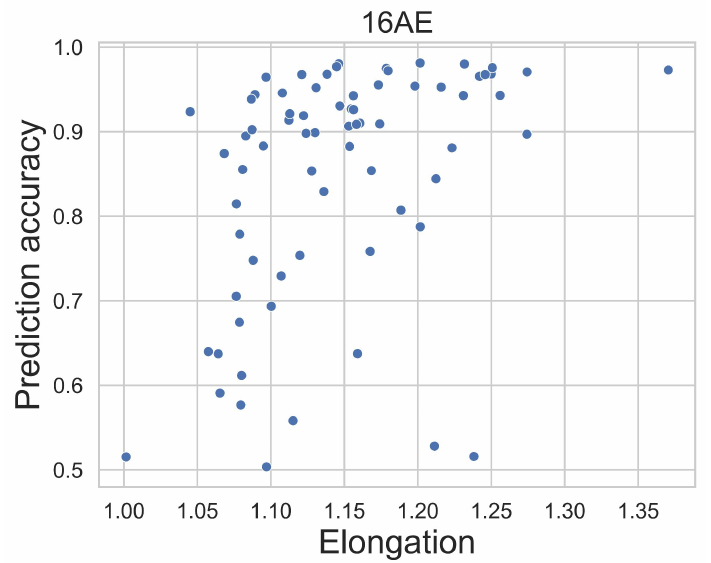}
\caption{Learning observed division patterns in \textit{A. thaliana} embryo at the 16C stage in the apical external domain (16AE): prediction accuracy as a function of mother cell elongation.}
\label{fig:sup-ambiguity16AE}
\end{figure}

\cleardoublepage

\begin{figure}
\includegraphics[width=\textwidth]{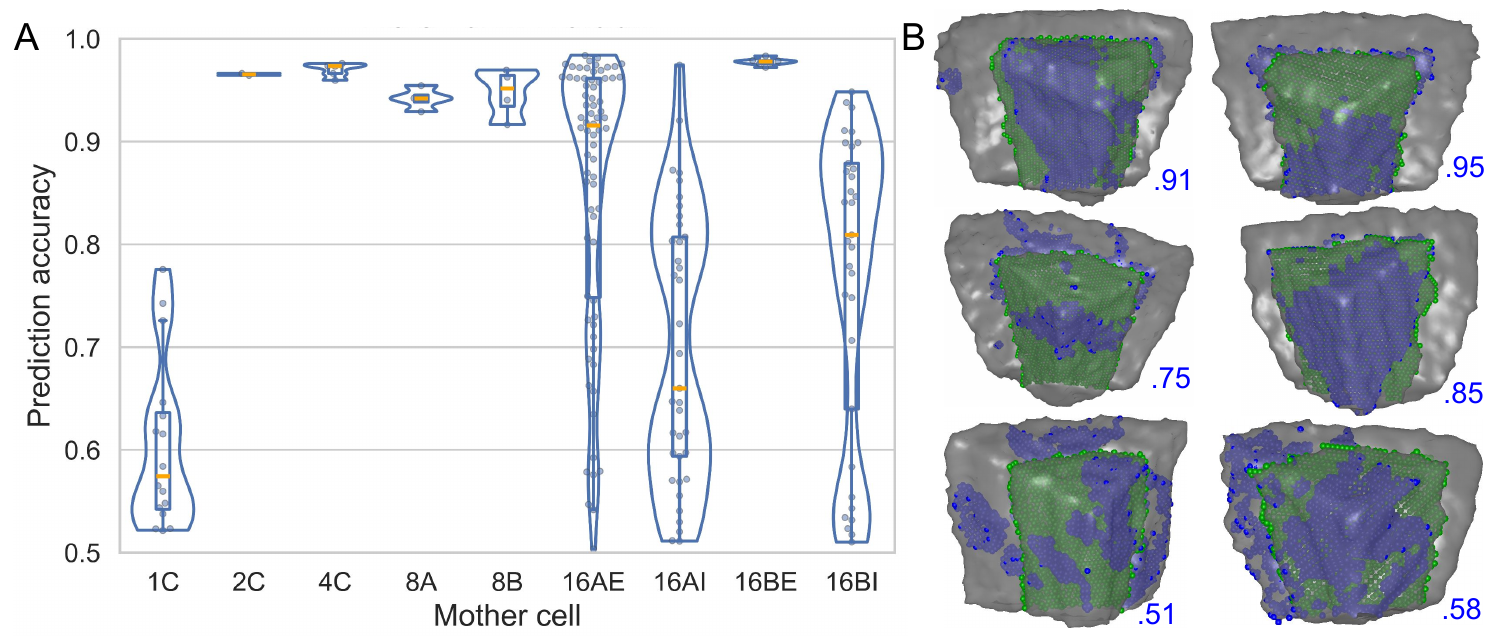}
\caption{Learning well-defined observed division patterns in \textit{A. thaliana} embryo: leave-one-out experiment on 16BI patterns. A model was trained on all well-defined divisions but those from the 16BI domain.
(A)~Accuracy distribution on the test sets. 
(B)~Predictions in the 16BI domain.
\textit{Green}: target (observed) division plane; \textit{Blue}: predicted plane. Numbers give the prediction accuracy.
}
\label{fig:sup-LOO16BI_Default}
\end{figure}

\cleardoublepage

\begin{figure}
\includegraphics[width=\textwidth]{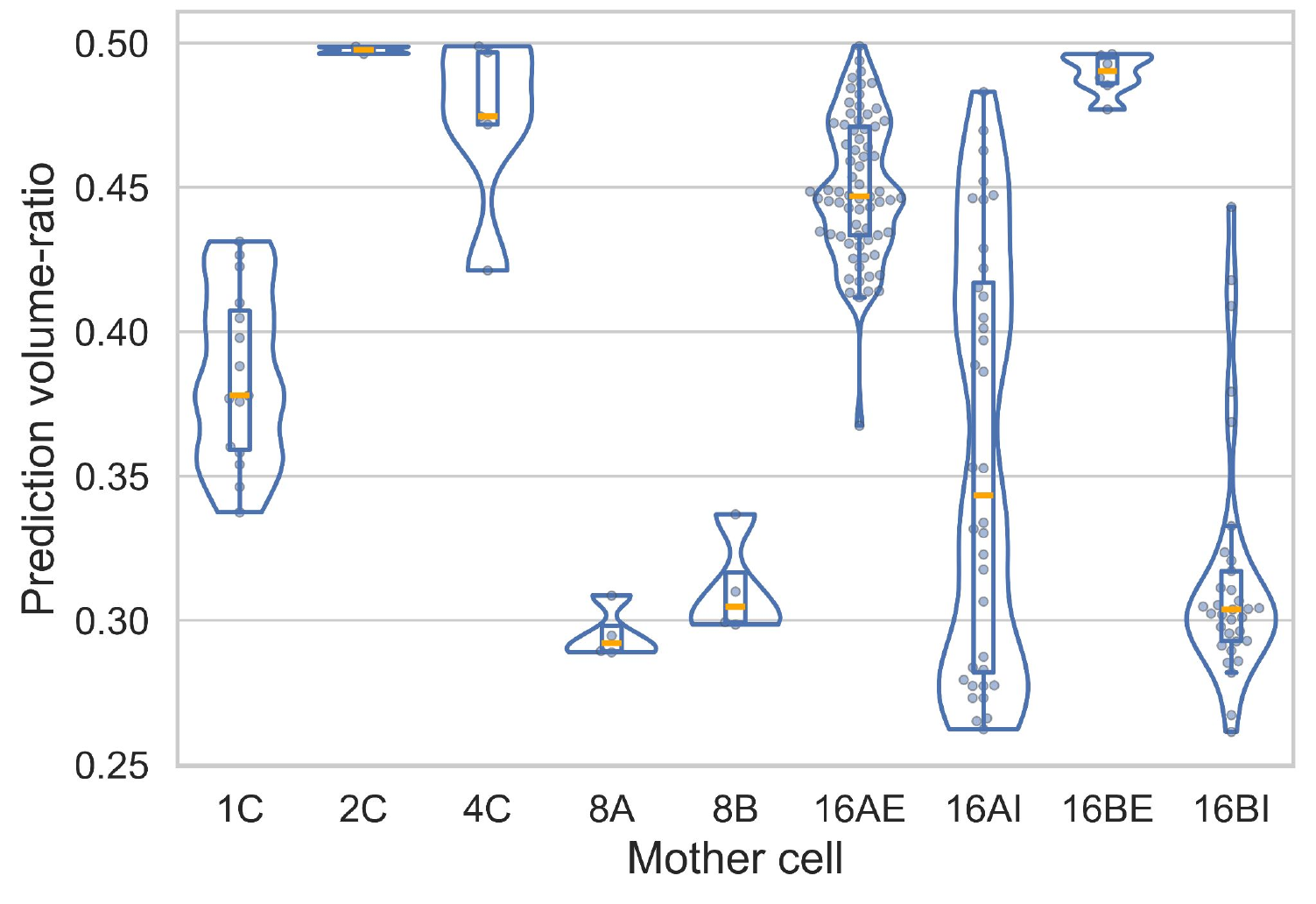}
\caption{Learning well-defined observed division patterns in \textit{Arabidopsis thaliana} embryo with volume standardization: predicted volume-ratios in a leave-one-out experiment on 16BI patterns. A model was trained on all well-defined divisions but those from the 16BI domain. All cell volumes were standardized during both training and inference (see Material and Methods).}
    \label{fig:sup-VR_LOO16BINorm}
\end{figure}

\end{document}